\documentclass[a4paper]{article}

\usepackage[margin=2.5cm]{geometry}
\usepackage[english]{babel}
\usepackage[utf8x]{inputenc}
\usepackage[T1]{fontenc}
\usepackage{ dsfont }
\usepackage{placeins}
\usepackage{multirow}
\usepackage{tabularx} 
\usepackage{lscape}
\usepackage{algpseudocode}
\usepackage{color,soul}

\usepackage{csvsimple}
\usepackage{longtable}
\usepackage{arydshln,amssymb,color}
\usepackage{multicol}
\usepackage{bm}
\usepackage{tikz}
\usetikzlibrary{arrows.meta}
\usepackage{amsmath}
\usepackage{graphicx}
\usepackage[colorinlistoftodos]{todonotes}
\usepackage{booktabs} 
\usepackage{xtab,booktabs}
\usepackage{footnote}
\makesavenoteenv{tabular}
\makesavenoteenv{table}
\usepackage{ dsfont }
\usepackage{ stmaryrd }
\usepackage{flushend}
\usepackage{natbib}
\usepackage{xspace}

\usepackage{subcaption}
\usepackage{animate}

\usepackage[linesnumbered,algoruled,boxed,lined,noend]{algorithm2e}
\SetKwInOut{KwParam}{Parameters}
\DeclareMathOperator*{\argmax}{argmax}

\usepackage{booktabs} 

\newtheorem{mydef}{Definition}

\newcommand\proposedmethodnamelong{Generalized Early Stopping for Direct Policy Search\xspace}
\newcommand\proposedmethodname{GESP\xspace}
\newcommand\nframeworksexperimentation{five\xspace}

\usepackage{xcolor}
\definecolor{myblue}{RGB}{0, 107, 179}
\definecolor{mygreen}{RGB}{0, 128, 0}
\usepackage[colorlinks=true,linkcolor=myblue,citecolor=mygreen,urlcolor=myblue,hyperfootnotes=false]{hyperref}
\hypersetup{
	linkbordercolor=myblue,
	citebordercolor=mygreen,
	urlbordercolor=myblue,
	pdfborderstyle={/S/U/W 1}
}

\title{Generalized Early Stopping in Evolutionary Direct Policy Search}
\author{Etor Arza$^1$ \and Leni K. Le Goff$^2$ \and Emma Hart$^2$}

\date{
	$^1$Basque Center for Applied Mathematics \\ \texttt{etorarza@gmail.com}\\%
	$^2$Edinburgh Napier University \\ \texttt{\{L.LeGoff2, E.Hart\}@napier.ac.uk}\\[2ex]%
}
\begin{document}
	\maketitle
	
	\begin{abstract}
		Lengthy evaluation times are common in many optimization problems such as direct policy search tasks, especially when they involve conducting evaluations in the physical world, e.g. in robotics applications.
		Often when evaluating solution over a fixed time period it becomes clear that the objective value will not increase with additional computation time (for example when a two wheeled robot continuously spins on the spot).
		In such cases, it makes sense to stop the evaluation early to save computation time.
		However, most approaches to stop the evaluation are problem specific and need to be specifically designed for the task at hand.
		Therefore, we propose an early stopping method for direct policy search.
		The proposed method only looks at the objective value at each time step and requires no problem specific knowledge.
		We test the introduced stopping criterion in \nframeworksexperimentation direct policy search environments drawn from games, robotics and classic control domains, and show that it can save up to $75\%$ of the computation time.
		We also compare it with problem specific stopping criteria and show that it performs comparably, while being more generally applicable.


	\end{abstract}


\newpage
 
	\section{Introduction}
	\label{sec:intro}

    Evolutionary algorithms (EAs) are increasingly being used in applications such as computer games \citep{desouzaBlindGameDesigner2014,hastingsEvolvingContentGalactic2009} and robotics \citep{hoffmannEvolutionaryAlgorithmsFuzzy2001,flemingEvolutionaryAlgorithmsControl2002} to learn control algorithms (policies), as well as being applied to classic control tasks such as the benchmark suites available in OpenAi Gym~\citep{brockmanOpenAIGym2016}.
    Often direct policy search algorithms such as EAs require a large number of evaluations: when these evaluations are costly in terms of time, this can result in extremely long learning times, which can be prohibitive in the worst case.
    Unfortunately many applications of interest suffer from this problem.
    For example, the protein folding problem~\citep{dillProteinFoldingProblem2008} requires costly simulations, while applications that involve a \textit{double} optimization process are also considered very computationally costly.
    This includes for example the joint optimization of robot morphology and control~\citep{hart2022artificial,legoffMorphoevolutionLearningUsing2021} in simulation (which typically use an outer loop to evolve body-plans and a nested inner-loop to evolve control), nested combinatorial optimization problems~\citep{wuSAFEScaleAdaptiveFitness2021,kobeaga2021solving} or hyperparameter optimization~\citep{desouzaCappingMethodsAutomatic2022}. 
    Specifically in robotics, evaluations that need to be conducted directly on a physical robot to avoid any reality-gap tend to be very time-consuming, while repeating lengthy evaluations also places considerable wear and tear on machinery, potentially leading to unreliable objective-function values.


	One approach to reducing the computational burden posed by expensive evaluation functions is to use a surrogate model~\citep{hwangFastpredictionSurrogateModel2018,ranftlBayesianSurrogateAnalysis2021}.
	Surrogate models try to replace the costly objective function with a cheaper alternative, that is usually less accurate but faster to compute~\citep{alizadehManagingComputationalComplexity2020}.
	This saves computation time because the number of function evaluations of the costly objective functions is reduced. However, selecting a suitable surrogate model can be challenging, typically involving the need to determine an appropriate trade-off between size (i.e. how much information is necessary to compute the surrogate model), the accuracy required, and computational effort (the time required for the surrogate modelling process itself)~\citep{alizadehManagingComputationalComplexity2020} which then influences the choice of surrogate model.

	Instead of reducing the number of evaluations, it is also possible to save computation time in these types of problems by stopping the evaluation of non promising solutions early.
	With this approach, given a fixed time budget in which to conduct evaluations which each have a maximum budget of $n$ seconds, it is possible to compute more evaluations than if every potential evaluation is run for exactly $n$ seconds.
	This is known as early stopping~\citep{li2017hyperband,hutter2019automated} or capping~\citep{hutter2009paramils,desouzaCappingMethodsAutomatic2022}. 
    Several early stopping approaches have been proposed for hyperparameter optimization, including \textit{irace}~\citep{lopez-ibanezIracePackageIterated2016,desouzaCappingMethodsAutomatic2022}, \textit{sequential halving}~\citep{karnin2013almost} and hyperband~\citep{li2017hyperband}.

    Early stopping has also been considered in the context of policy search.
    For example, when learning to control a robot in a simulation,  if the robot gets stuck (it does not move) it is useful to stop the evaluation early~\citep{legoffMorphoevolutionLearningUsing2021}.
    However, there are two limitations associated with these problem specific methods.
    First, these approaches, in some cases, can fail to stop the evaluation even if it is clear that additional time is not going to improve the objective value.
    For example, when a two wheeled robot continuously spins on the spot, it would still register as moving, but it is completely useless to continue evaluating.
    Secondly, they require problem specific knowledge, such as detecting when the robot is not moving, which might not always be trivial (for example when dealing with robots in the real world).

    The main contribution of this paper is to show that generic early stopping is applicable to direct policy search via evolutionary algorithms.
    Similar to early stopping methods for hyperparameter optimization~\citep{desouzaCappingMethodsAutomatic2022,li2017hyperband,hutter2019automated,karnin2013almost} and unlike current early stopping criteria for direct policy search, the proposed approach only needs the objective value to decide when to stop the evaluation of the robots.
    We demonstrate both the efficacy and generality of the method in a wide-ranging experimental section in \nframeworksexperimentation different direct policy search environments, showing that the proposed approach significantly reduces the optimization time of direct policy search algorithms in a wide variety of control tasks.

    The paper is organized as follows. We first discuss some related work to position the proposed method in the literature. In the next section, we provide a formal definition of the problem and introduce the proposed early stopping method. In Section~\ref{section:experimentation} we present the \nframeworksexperimentation part experimental study on the applicability of the proposed method in direct policy search tasks. Finally, Section~\ref{section:conclusion} concludes the article.

\section{Related work}

    Many direct policy search tasks in the literature use problem specific early stopping methods to save computation time.
    One of the earliest examples is probably the \textit{cart pole} control problem~\citep{bartoNeuronlikeAdaptiveElements1983}.
    In this problem, a cart needs to balance a pole by moving left or right (see the animation on \href{https://web.archive.org/web/20230324042842/https://www.gymlibrary.dev/_images/cart_pole.gif}{OpenAI's website}\footnote{https://www.gymlibrary.dev/\_images/cart\_pole.gif}).
    In the \href{https://web.archive.org/web/20221231160811/http://incompleteideas.net/sutton/book/code/pole.c}{original implementation}\footnote{http://incompleteideas.net/sutton/book/code/pole.c} by \citet{suttonTemporalAspectsCredit1984}, the evaluation is usually stopped after $10^5$ steps (episode length), but is terminated early when the pole is considered not balanced for 100 steps.
    In the \href{https://web.archive.org/web/20221207031552/https://www.gymlibrary.dev/environments/classic_control/cart_pole/}{modern version}\footnote{https://www.gymlibrary.dev/environments/classic\_control/cart\_pole/} of the same problem by OpenAI, the episode length is $500$ and the evaluation is terminated early when the pole is not considered balanced, or the cart has moved too much to one of the sides.

    Another more modern example with early stopping is the \href{https://web.archive.org/web/20221211213854/https://www.gymlibrary.dev/environments/mujoco/ant/}{Ant control task}\footnote{https://www.gymlibrary.dev/environments/mujoco/ant/} in the MuJoCo environment in OpenAI gym~\citep{brockmanOpenAIGym2016}.
    In this task, the control of an ant needs to be learned, such that the traveled distance is maximized, while minimizing the energy expenditure and the contact force (see the \href{https://web.archive.org/web/20221214135341/https://www.gymlibrary.dev/_images/ant.gif}{animation}\footnote{https://www.gymlibrary.dev/\_images/ant.gif}).
    In this task, the episode length is $1000$ steps, but an evaluation is stopped early if the vertical position of the torso is not in the interval $[0.2, 1.0]$, or if there are numerical errors in the simulator.

    A limitation of the early stopping criteria of these and other direct policy search tasks is that they are highly problem specific.
    They need to be carefully designed, taking into account the problem at hand. 
    For example, in the previous \textit{ant} example, choosing to stop the evaluation based on the position of the torso is not trivial, and requires understanding of what is a desirable position of the torso.

    There has been plenty of work in early stopping based on the objective function alone, but most of it has focused on hyperparameter optimization.
    Some of the best known early stopping approaches for hyperparameter optimization are the \textit{irace} algorithm~\citep{lopez-ibanezIracePackageIterated2016,desouzaCappingMethodsAutomatic2022}, the sequential halving algorithm~\citep{karnin2013almost,hutter2019automated} and hyperband~\citep{li2017hyperband,pmlr-v80-falkner18a,zimmerAutoPytorchMultiFidelityMetaLearning2021}.
    Early stopping approaches work very well in hyperparameter optimization because the evaluation of solutions can be paused and resumed.
    Consequently, it is possible to evaluate a set of solutions simultaneously and compare their partial objective values with one other, discarding poorly performing candidates before continuing the evaluations.
    This is not always possible in direct policy search tasks, especially those that run in the real-world.
    For example, it would not make sense to pause the evaluation of a controller in a real-world robot, evaluate a different controller in the same robot, and then resume the previous evaluation. 
    The same applies to simulation, where it is not trivial to implement a way to save the state of the simulation and load it later.

    More recently, \citet{desouzaCappingMethodsAutomatic2022} proposed a set of early stopping methods that only take into account the objective function.
    The early stopping methods proposed by \citet{desouzaCappingMethodsAutomatic2022} record the objective values of previous solutions during each time step.
    Then, when new solutions are evaluated, it is possible to stop their evaluation early at time step $t$ if the solution is expected to perform poorly with additional computation time.
	\citet{desouzaCappingMethodsAutomatic2022} propose two types of methods: profile envelopes and area envelopes.
	A profile envelope is a reference objective function $f^{*}(t)$ that serves as a stopping criterion for time step $t$: assuming a maximization problem with a positively defined monotone increasing objective function $f$, $\theta$'s evaluation is stopped at time step $t$ if $f^{*}(t) > f[t](\theta)$.
    The area envelope, on the other hand, is a stopping criterion that takes into account the whole trajectory.
    Given a maximum performance area $A$---a positive real value---the evaluation of $\theta$ is stopped at time step $t$ if $\int_{x=0}^t f[x](\theta)dx < A$.
	
	Although it might be possible to adapt \citet{desouzaCappingMethodsAutomatic2022}'s methods for direct policy learning, they are not directly applicable, as they were designed specifically for hyperparameter optimization.
	In this regard, their method assumes a monotone decreasing objective value which is not always true for policy learning problems.
	In addition, \cite{desouzaCappingMethodsAutomatic2022}'s method assumes that partially evaluated solutions are eligible, which is true for hyperparameter optimization but not for direct policy learning.

	Early Stopping has also been considered in the field of multi-fidelity optimization in the context of airfoil design optimization via computational fluid dynamics.
	\citet{forrester2006optimization} proposed using early stopping to build more accurate surrogate models.
	Specifically, given a fixed computation budget, a more accurate surrogate model can be obtained by evaluating more solutions for less time each (via early stopping).
	When to stop the evaluation of the solution is decided by analyzing the convergence of the surrogate with a previous version that was computed with less computation time.
	\citet{pichenyNonstationarySpacetimeGaussian2013} further refined the method such that not all solutions need to be evaluated with the same computation time, and solutions that are expected to perform poorly can be stopped earlier.
	To achieve this, a Gaussian process is fitted that jointly models the design parameter space and computational time.

	By making certain assumptions on the objective function and considering \textit{$1 + \lambda$ evolution strategies}, \citet{bongard2010utility,bongardInnocentProvenGuilty2011} proposed an early stopping mechanism for multi-objective evolution of robots, based on the objective function.
	The approach involves stopping the evaluation of candidates once it is impossible for them to beat the best found candidate in the current generation.
	This early stopping approach has the very good property of not changing the final outcome while still saving computation time.
	However, the applicability of Bongard's method in the general case is very limited, as it depends on both a specific definition of the objective function and the use of the \textit{$1 + \lambda$ evolution strategies} algorithm.\footnote{The objective function to be maximized needs to be a combination of other sub-objective functions in different sub-tasks, assuming that the sub-objective values are always negative. 
    With this objective function, it is not necessary to evaluate the solution in every sub-task, as the objective value can only decrease with further evaluations. 
    Consequently, once the objective value of a solution is lower than the best candidate in the current population, the evaluation can be stopped, as it is guaranteed that it will not outperform the best candidate.
    When considering this early stopping criterion in combination with the use of the \textit{$1 + \lambda$ evolution strategies} algorithm, the same final solution is obtained while saving computation time.}

	\citet{wangSubgoalbasedExplorationBayesian2022} also considered an early stopping approach on an exploration environment.
	Their approach involves using Bayesian Optimization to efficiently choose the maximum episode length and other relevant hyperparameters during training, obtaining a speedup over the default training procedure.
	However, as with the approach by \citet{bongard2010utility}, the approach by \citet{wangSubgoalbasedExplorationBayesian2022} is specific to the environment.

    In summary, the early stopping approaches that have been tested in direct policy search tasks either require problem specific knowledge, or a specific definition of the objective function and learning algorithm.
    This motivates our proposal which addresses both these issues.

	
%
%
%
%

	\section{\proposedmethodnamelong (\proposedmethodname)}
	
    We propose a simple early stopping method for direct policy search tasks that overcomes the limitations discussed in the previous section.
    The proposed method is general and makes no assumptions about the optimization problem or the objective function or the learning algorithm.
    It uses the output of the objective function without the need of problem specific information.
    Specifically, the proposed approach is applicable to the optimization problem defined as follows:

	\begin{mydef}
		\label{def:estimable_optimization_problem}
		Let $T$ be a maximum computation time budget, $f$ an objective function and $\Theta$ a solution space.
		We define a optimization problem $\argmax_{\theta \in \Theta} f(\theta)$ with these five additional properties:
		\begin{enumerate}
			\item Computing $f(\theta)$ has a time cost $t_{max}$.
			\item No further evaluations are possible once the computation budget $T$ is spent. 
			\item Instead of computing $f(\theta)$ exactly, it can be approximated for any lower time cost $t < t_{max}$. We denote the approximation in time $t$ as $f[t](\theta)$. 
			\item The approximation with $t_{max}$ is the original objective value: $f(\theta) = f[t_{max}](\theta)$.
			\item Approximate solutions with $t < t_{max}$ are not eligible to be chosen as the best.
		\end{enumerate}
	\end{mydef}

	In addition, \proposedmethodname is specifically designed for problems with these three additional properties. 

	\vspace{0.5em}
	\textbf{Property 1: Incremental approximation (early stopping possible)}
	
	Given a solution $\theta$, if it has been approximated with time $t$, it is possible to resume the evaluation until time $t + t_\delta$. This extension has time cost $t_\delta$. In simpler terms, we can choose whether to stop the evaluation at time step $t$ or resume evaluation, after observing $f[t](\theta)$.

	Note that this property does not hold for all the problems in which an approximate objective function exists.
	For example, the objective value on the wind turbine design problem by \citet{zarketa-astigarragaComputationallyEfficientGaBased2023} can be approximated by reducing the number of divisions on the blades for the CDF calculation.
	The lower the number of divisions, the less time it takes to compute the objective value, but the approximation is also less accurate.
	However, the number of subdivisions needs to be chosen \textit{before} the objective value is computed.
	Consequently, it is not possible to resume the evaluation or increase the number of divisions without starting the evaluation again from scratch.

	For problems like this in which the objective function cannot be incrementally computed, other time reducing methods are possible.
	For example, \citet{echevarrieta2024speeding} proposed a method that can find the optimal cost for these types of problems.
	In the problem above, the method compares the rankings of objective values with different numbers of divisions, and chooses the lowest (most efficient) number of divisions that still ranks the solutions like the original objective function.

	The policy learning problems and environments considered on this paper, on the other hand, can all be early stopped.
	Reinforcement learning tasks in general can be stopped at each time step, as they are modeled as a Markov Decision Process.
	Agents observe the state, interact with the environment with an action, and receive a reward.
	This cycle is repeated many times and can be interrupted every time the reward is observed.

	\vspace{0.5em}
	\textbf{Property 2: Resuming an evaluation is not possible}
	
	If the evaluation of a solution has already been early stopped and we started evaluating another solution, then it is not possible to extend the evaluation of the previous solution.
	Well known early stopping methods from the literature such as sequential halving~\citep{karnin2013almost} or hyperband~\citep{li2017hyperband} are not applicable to problems where it is not possible to resume a previous evaluation.
	In contrast, both \citet{desouzaCappingMethodsAutomatic2022}'s method and \proposedmethodname are applicable on problems with this property (although they can also be applied in problems that do not have this property).

	The limitation of being \textit{unable to resume the evaluation} is very relevant in policy learning tasks, specially, when they are carried out in the real world.
	For example, when carrying out direct policy search in real robots, it does not make sense to evaluate policy $A$ for $2$ seconds, immediately change to another policy $B$ for $3$ seconds and then decide to go back to continue with the evaluation of policy $A$.
	Restarting the evaluation of a robot takes time, as there is usually only \textit{one} robot.
	The robot needs to be reset to the initial state (often with human intervention) before another policy can be evaluated.
	In addition, resuming the evaluation from a partially evaluated policy requires resetting the robot to the state in which the evaluation was stopped, which is impossible in many cases.

	\vspace{0.5em}
	\textbf{Property 3: Approximation quality}
	
	Given any two solutions $\theta_1,\theta_2$, \proposedmethodname assumes that the approximation of the objective function, is \textit{usually} able to properly identify the best of these solutions.
	Hence, if $f(\theta_1) \geq f(\theta_2)$ then the probability of $f[t](\theta_1) \geq f[t](\theta_2)$ should be high, specially when $t$ is close to $t_{max}$.

	In other words, \proposedmethodname and the rest of the early stopping methods in the literature assume that if a solution performs poorly when approximated, then its objective value is also expected to be low.
	This is an assumption that reasonably holds for hyperparameter optimization~\citep{li2017hyperband}, and is required by all early stopping approaches.
	In the context of policy learning, this property implies that the cumulative reward does not drastically change in a few steps.
	Instead, Property~3 implies that the cumulative reward increases or decreases slowly throughout the episode.


  \paragraph{Generalized Early Stopping for Direct Policy S.earch}
	Assuming a maximization optimization problem, \proposedmethodname involves stopping the evaluation of a solution at time step $t>t_{grace}$ when certain conditions are met.
    Specifically, we stop the evaluation of a solution $\theta$ at time step $t$ if  Equation~\eqref{eq:early_stopping_generalized} below is satisfied.
	
	\begin{equation}
		\label{eq:early_stopping_generalized}	
		\max\{f[t](\theta), f[t-t_{grace}](\theta) \} < \min\{f[t](\theta_{best}), f[t-t_{grace}](\theta_{best})\}
	\end{equation}

    The grace period parameter $t_{grace}$ is a parameter that establishes a minimum time for which all candidates will be initially evaluated regardless of their objective value.
	In addition, it determines the bonus evaluation time given to new candidate solutions.
    The evaluation of the new candidate solution $\theta$ is stopped when its objective value at time step $t$ is worse than the objective value of the best solution $\theta_{best}$ at time $t - t_{grace}$.
	This gives new candidates $t_{grace}$ extra time steps to achieve the level of performance of the current best solution

	In Algorithm~\ref{algo:early_stopping} we show the pseudocode of \proposedmethodname.
	First we initialize the reference objective values to $-\infty$ (lines 1-2).
	Then, starting with the first time step $t=1$, we evaluate the approximate objective function $f[t](\theta)$ at that time step (line 4).
	Then, if $t > t_{grace}$ and Equation~\eqref{eq:early_stopping_generalized} are satisfied (line 5), the objective function of $\theta$ is approximated as $f[t](\theta)$ (line 6), $\theta$ is not evaluated in time step $t+1$ and beyond, and the reference objective values are not updated.
	Otherwise, we evaluate $\theta$ at time step $t+1$.
	Finally, if $\theta$ makes it to time step $t_{max}$ and a new best objective value is found  $f[t_{max}](\theta_{i}) > f[t_{max}](\theta_{best})$, we replace the reference objective values with the new ones (lines 7-8).
%

	\begin{algorithm}[t]
		\begin{small}
			\DontPrintSemicolon 
			\caption{Evaluate solution with early stopping}
			\label{algo:early_stopping}
			\KwIn{\ \\%
				$f[t](\cdot)$: The approximation of the objective function with time $t$. $f[t_{max}](\cdot)$ is the objective function.\\
				$\theta$: The solution to be evaluated.\\
				$t_{grace}$: The grace period parameter. \\
				$t_{max} $: The maximum evaluation time.
			}%

			\SetAlgoLined
			\vspace{0.25cm}
			\If{$f[t](\theta_{best}) = \phi$}
			{	

				$f[t](\theta_{best}) \gets - \infty, \ \ \ \forall  t = 1,...,t_{max}$\; 
			}
			
			\For{$t=0,...,t_{max}$}{
				compute $f[t](\theta)$\;
				\If{$t > t_{grace}$  \normalfont\textbf{and} Equation~\eqref{eq:early_stopping_generalized} is satisfied}
				{
					\textbf{return} $f[t](\theta)$\;
				}

			}
			\If{$f[t_{max}](\theta) > f[t_{max}](\theta_{best})$}
			{
				$f[t](\theta_{best}) \gets f[t](\theta), \ \ \ \forall  t = 1,...,t_{max}$\; 
			}


			\textbf{return} $f[t_{max}](\theta)$\;
			
		\end{small}    
	\end{algorithm}

    Our approach is similar to \citet{desouzaCappingMethodsAutomatic2022}'s early stopping methods for hyperparameter optimization.
    \citet{desouzaCappingMethodsAutomatic2022}'s approach generates a stopping criterion based on a set of already evaluated solutions, such that the evaluation of new candidate solutions can be stopped early when they perform relatively poorly.
    However, \citet{desouzaCappingMethodsAutomatic2022}'s approach is not directly applicable to direct policy search, as it assumes a monotone objective function as well as the eligibility of partially evaluated solutions.
    In the following, we discuss a few details that need to be taken into account when proposing a early stopping method for direct policy search (in contrast to hyperparameter optimization).

    \subsection{Applicability on direct policy search}
    \label{sec:overcoming_monotone_increasing}

    Early stopping through the objective function alone usually assumes a monotone increasing objective function, which is true for hyperparameter tuning.
    However, this assumption does not hold for direct policy search.
    Specifically, early stopping without a monotone increasing objective function creates two issues: and we propose two possible modifications of \proposedmethodname to overcome them.

    We motivate and explain these two issues with an example.
    Let us consider the reinforcement learning \href{https://web.archive.org/web/20220903205443/https://www.gymlibrary.dev/environments/classic_control/pendulum/}{pendulum task}\footnote{https://www.gymlibrary.dev/environments/classic\_control/pendulum/}.
    The reward in this task is inversely proportional to the speed and the angle of the pendulum (assuming the angle at the upright position is 0).  
    The goal in this task is to maximize the sum of all the rewards in $t_{max}$ time steps, where the reward in each step is in the interval $(0, -16.27)$.
    This means that in each time step, the objective function can only be lower than in the previous time step (the objective function is monotone decreasing).
    Consequently, if early stopping is applied in this problem, the solutions that are stopped earlier will have a better objective function than if they had been evaluated for the maximum time $t_{max}$.

    Issue (1) involves correctly reporting the best found solution.
    Unless we assume a monotone increasing objective function, the best found solution might be a partially evaluated solution.
    Consequently, if the issue is not addressed, the reported objective value could potentially be better than the actual objective value.

	For example, in the \textit{pendulum} task, if the pendulum is initialized in the downwards position and a policy $\theta_{worst}$ applies no torque, then, then the reward is -16.27 (the worst possible reward) in each time step ($f[t](\theta_{worst}) = -16.27 \cdot t$).
	However, when applying \proposedmethodname,	at time step $t = t_{grace} + \delta$, the policy is likely to be early stopped.
	Recall that the evaluation is stopped if

	\begin{equation}
	\max\{f[t](\theta_{worst}), f[t-t_{grace}](\theta_{worst}) \} < \min\{f[t](\theta_{best}), f[t-t_{grace}](\theta_{best})\}
	\end{equation}
	
	considering that the reward in each time step is defined on the interval $(0, -16.27)$, $t = t_{grace} + \delta$, and $f[\delta](\theta_{worst}) = -16.27 \cdot \delta $, we simplify the above equation to

	\begin{equation}
	-16.27 \cdot \delta  < f[t_{grace} + \delta](\theta_{best})
	\end{equation}

	Now, for the sake of simplicity, lets assume that the best policy $\theta_{best}$ so far is able to control the pole on the first $t$ time steps such that the reward is $-5$ on average.
	Then, the above equation would be simplified to 

	\begin{equation}
	f[\delta](\theta_{worst}) = -16.27 \cdot \delta < -5 \cdot (t_{grace} + \delta)
	\end{equation}

	\begin{equation}
	0.443656 \cdot t_{grace} < \delta
	\end{equation}

	Thus, at time step $t = 2\cdot t_{grace}$ the evaluation will have already stopped, which will give the policy $\theta_{worst}$ an objective value of $-32.54 \cdot t_{grace}$ (or better).
	Now, if $t_{grace}$ is set to $0.05 \cdot t_{max}$, then the objective value of $\theta_{best}$ will be $-5 \cdot t_{max} = -100 \cdot t_{grace}$.
	Thus, the reported objective value of the best found solution $\theta_{best}$ will be more than 3 times worse than for the worst possible policy $\theta_{worst}$.
	Even though $\theta_{worst}$ is a terrible policy, it triggers the early stopping very quickly, and obtains a better objective value than a policy that performs well and was evaluated for the entire episode.
	Hence, without taking into account Issue (1), the best reported solution could be one that immediately triggers the early stopping criterion.
	Overcoming this issue is simple with Modification (1): do not update the best found solution \textit{unless} the new best solution has been evaluated for the maximum time $t_{max}$.

    Issue (2) is not as critical as Issue (1), and is related to the credit assignment during the optimization process.
    Although Issue (2) does not produce an incorrect result per se, it might potentially set back the optimization.
    In monotone increasing problems (such as hyperparameter optimization), a poor solution that is early stopped might get a worse than deserved objective value, as with additional evaluation time it might have been able to increase its objective value.
    This is usually not considered a problem, as it is not expected that the learning algorithm is impacted in a negative way.
    Basically early stopping in this case favors solutions that quickly converge towards good objective values over those that are slow to converge.
    However, in problems with decreasing objective functions (such as the \textit{pendulum} task), poor solutions might be assigned an objective value (due to early stopping) that is in fact better than the objective value that would have been obtained if the evaluation was continued for the maximum evaluation time. 
    This might pose a challenge for the optimization algorithm.

    For example, in the \textit{pendulum} task, if a policy applies no torque then the pendulum does not move and the evaluation is stopped early.
    This is a very poor policy, but since it triggers the early stopping very quickly, it obtains a better objective value than a policy that is able to correctly balance the pole and is evaluated for the entire episode.
    This is because in the \textit{pendulum} task, the reward is negative in each time step, and the total reward of the policy that optimally balances the pole has many time steps with a negative reward (during which the pole is being moved towards the balancing point).
    The learning algorithm might therefore optimize the policy to trigger the early stopping as quickly as possible, which is obviously undesirable.

    It is possible to overcome Issue (2) by modifying the objective function.
    It is enough to redefine the objective function such that it is monotone increasing.
    To achieve this, we can add a constant value $k$ to the objective function in each time step, ensuring that the redefined objective function is monotone increasing. For instance, in the \href{https://web.archive.org/web/20220903205443/https://www.gymlibrary.dev/environments/classic_control/pendulum/}{pendulum task} (with a reward in the interval $(0, -16.27)$ in each time step), it is sufficient to redefine the objective function as $f_{new}[t](\theta) = f[t](\theta) + t \cdot 16.28$.
    By adding Modification (2), we also overcome Issue (1).

    However, in this study, we deliberately chose not to redefine the objective function (we do not apply Modification (2)).
    We propose \proposedmethodname as a plug and play method that is compatible with as many problems as possible and requires no modifications in the objective function and can work alongside other problem specific stopping criteria.
    The purpose of the experimentation in this work is to showcase the benefit of applying \proposedmethodname on direct policy learning environments.
    In this sense, modifying the objective functions of the environments would obscure the contribution of \proposedmethodname.

    Even with only Modification (1), we show that \proposedmethodname rarely decreases the performance and significantly speeds up the search process in a wide variety of tasks (despite Issue (2)), even in tasks with monotone decreasing objective functions such as \textit{pendulum} (Section~\ref{sec:classic_control}).

	\section{Experimentation}
	\label{section:experimentation}

	To validate the proposed approach, we run several experiments in different direct policy search tasks with different evolutionary learning algorithms to demonstrate the  benefit of using \proposedmethodname. 
    We chose different tasks with different learning algorithms to show that \proposedmethodname is applicable in different scenarios and across a range of evolutionary algorithms.
    In some of these tasks, the original authors have proposed a problem specific stopping criterion to save computation time, in which case, we also compare \proposedmethodname to the problem specific approach.
    The experimentation is carried out in \nframeworksexperimentation different environments, each with different properties and learning algorithms.
    Table~\ref{tab:summary_environments} lists the environments considered.

    \begin{table}
    	\centering
    	\resizebox{\linewidth}{!}{%
    		\begin{tabular}{>{\centering\hspace{0pt}}m{0.13\linewidth}>{\centering\hspace{0pt}}m{0.15\linewidth}>{\centering\hspace{0pt}}m{0.4\linewidth}>{\hspace{0pt}}m{0.45\linewidth}}
    			Name                          & Environment                     & Learning \par{}algorithm                    & Task                                                                                                                             \\ \hline
    			\textbf{classic control}      & Classic control                 & CMA-ES  \citep{igelComputationalEfficientCovariance2006}                                     & \textit{cart pole} and \textit{pendulum}.                                                                                         \\ \hline
                \textbf{super mario}          & Super Mario                     & NEAT   \citep{stanleyEvolvingNeuralNetworks2002}                                     & Move the character ``Mario'' to the right as much as possible.                                                              \\ \hline
    			\textbf{mujoco}               & Mujo-co                         & CMA-ES    \citep{igelComputationalEfficientCovariance2006}                                  & \textit{half cheetah}, \textit{inverted double pendulum}, \textit{swimmer}, \textit{ant}, \textit{hopper} and \textit{walker2d}. \\ \hline
                \textbf{NIPES explore}        & 8 x 8 grid \par{}with obstacles & NIPES    \citep{legoffSampleTimeEfficient2020}                                   & Move a~ robot with four wheels and 
                two sensors and visit as many squares possible in $30$ seconds.                               \\ \hline
    			\textbf{~~ L\nobreakdash-System ~~} & Based on\par{}OpenAI gym        & $\lambda$~+ $\mu$ ES with L-System encoding \newline \citep{veenstra2020different} & Learn the morphology and control of virtual creatures and move to the right as much as possible.                                 \\ \hline
    		\end{tabular}
    	}
    	\caption{Environments in the experimentation}
    	\label{tab:summary_environments}
    \end{table}

	\subsection*{Methodology}

	For each experiment, we record the best objective value found so far with respect to the computation time (known as the \textit{attainment trajectory}~\citep{dreo2021extensible}).
	Each experiment is repeated 30 times, and the median and interquantile ranges are reported.
	When comparing two attainment trajectories, we perform a pointwise two sided Mann-Whitney test \citep{10.2307/2236101, arzaComparingTwoSamples2022} with $\alpha = 0.01$.
	The test is performed pointwise with no familywise error correction~\citep{korpelaConfidenceBandsTime2014}.

	When performing no correction, on average (if all the experiments were repeated $\infty$ times), the proportion of the points in which the test is falsely rejected is 0.01.
	We think that falsely rejecting the null-hypothesis (finding a difference when in reality there is none) in a small proportion of the optimization procedure is acceptable~\citep{hardle2004nonparametric}.

	Consequently, we expect that in a small proportion of all the points on the x-axis, the statistical test is falsely rejected (difference is found when in reality there is none).
	As a consequence, in the results figures in the paper, on average\footnote{As with every hypothesis test (frequentist) approach, \textit{on average} in this context means if we were to repeat this same experiment many times~\citep{conover1980practical}.} we can expect to observe a false statistically significant difference in $1\%$ of the length.

We also added the best \textit{found} values (the best value observed during our executions) instead of the best \textit{known} values (best values observed in the literature) as a reference.
Two of the benchmarks (\textbf{super mario} and \textbf{L\nobreakdash-System}) are not very well known, and do not have best \textit{known} values available.
Even for the tasks where the best \textit{known} value is available, it depends on the type of optimization algorithm considered.
Specifically, in this work, we present a technique (\proposedmethodname) that can speed-up \textbf{existing} direct policy search algorithms.
Therefore, the purpose of the experimentation is to demonstrate the benefit of applying \proposedmethodname to direct policy search algorithms (we are not trying to introduce a state of the art policy learning algorithm).
In many of the problems, the best known value in the literature is very far from what can be achieved with direct policy search.
For example, the results available on \href{https://spinningup.openai.com/en/latest/spinningup/bench.html}{OpenAI's spinning up benchmark} (\href{https://web.archive.org/web/20230602053435/https://spinningup.openai.com/en/latest/spinningup/bench.html}{also archived}) an objective value of almost 15000 can be reached with the \href{https://spinningup.openai.com/en/latest/algorithms/sac.html}{soft actor critic algorithm}, while the best \textit{found} with the direct policy search method for this task in our work is around 3000.

	\subsection{Classic control tasks (classic control)}
    \label{sec:classic_control}
 
    OpenAI Gym~\citep{brockmanOpenAIGym2016} is a framework to study reinforcement learning.
    Among the environments available in OpenAI Gym, we have the classic control tasks \textit{cart pole}~\citep{bartoNeuronlikeAdaptiveElements1983} and \href{https://web.archive.org/web/20220903205443/https://www.gymlibrary.dev/environments/classic_control/pendulum/}{\textit{pendulum}}.
    In the \href{https://github.com/rlworkgroup/garage}{garage}\footnote{https://www.gymlibrary.dev/environments/classic\_control/pendulum/} framework, there is an \href{https://github.com/rlworkgroup/garage/blob/2d594803636e341660cab0e81343abbe9a325353/src/garage/examples/np/cma_es_cartpole.py#L4}{example script}\footnote{https://github.com/rlworkgroup/garage/blob/2d594803636e341660cab0e81343abbe9a325353/ src/garage/examples/np/cma\_es\_cartpole.py\#L4} (also \href{https://web.archive.org/web/20230529153930/https://github.com/rlworkgroup/garage/blob/2d594803636e341660cab0e81343abbe9a325353/src/garage/examples/np/cma_es_cartpole.py}{archived}) that uses CMA-ES to learn the policy for the \textit{cart pole} task.
    The same learning algorithm was considered for \textit{pendulum}.
    We set the grace period parameter to 20\% of the maximum time: $t_{grace} = 0.2 \cdot t_{max}$.

    The objective function in \textit{cart pole} is the number of timesteps before it is terminated because out of bounds or because the pole is no longer in the upright position (the reward is $1$ in every time step).
    Consequently, any approximation of the objective function that has not yet been terminated is $f_{cartpole}[t](\theta) = t$.
    As mentioned before, pendulum has a monotone decreasing objective function with a reward in the interval $(0, -16.27)$ in each time step.
    The policies are learned with CMA-ES.

    Let us first consider the \textit{cart pole} experiment.
    Due to the definition of the objective function, applying \proposedmethodname has no effect in this case: the condition in Equation~\eqref{eq:early_stopping_generalized} will never be satisfied, and no early stopping will happen.
    Consequently, we expect that there is no difference experimentally between applying \proposedmethodname and applying no early stopping (denoted as ``Standard'' in the figures in the paper).
    In fact, since applying \proposedmethodname has no effect, the null hypothesis is true for this task.
 
\begin{figure}[h]

		\centering %

		\begin{subfigure}{0.45\textwidth}
			\includegraphics[width=\linewidth]{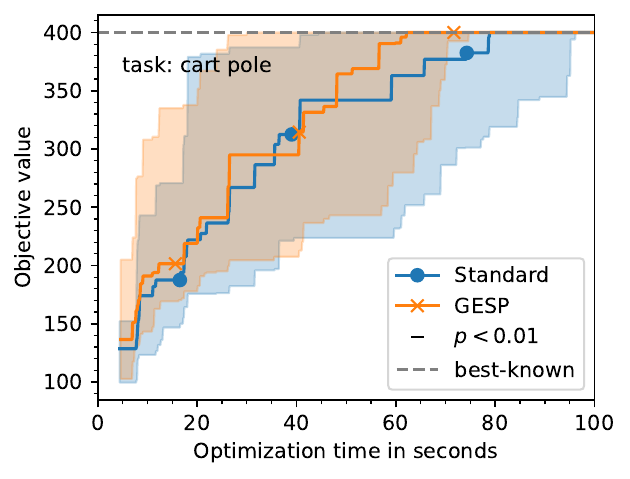}
			\caption{\textit{cart pole}}
			\label{fig:gymenvnamecartpole-v1expline}
		\end{subfigure}\hfil 
		\begin{subfigure}{0.45\textwidth}
			\includegraphics[width=\linewidth]{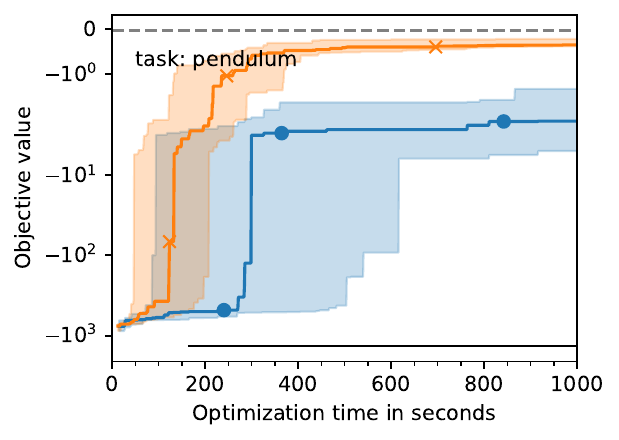}
			\caption{\textit{pendulum}}
			\label{fig:gymenvnamependulum-v1expline}
		\end{subfigure}\hfil 
		
		\caption{The objective value of the agents with and without \proposedmethodname with respect to computation time (\textbf{classic control}).}
		
	\end{figure}

    The result for the \textit{cart pole} experiment shown in Figure~\ref{fig:gymenvnamecartpole-v1expline} confirms this experimentally.
    Visually, there is no difference between the two stopping criteria, also suggested by the lack of statistical significant difference of the two sided Mann-Whitney test at $\alpha = 0.01$.

\begin{figure}
        \centering
		\includegraphics[width=0.45\linewidth]{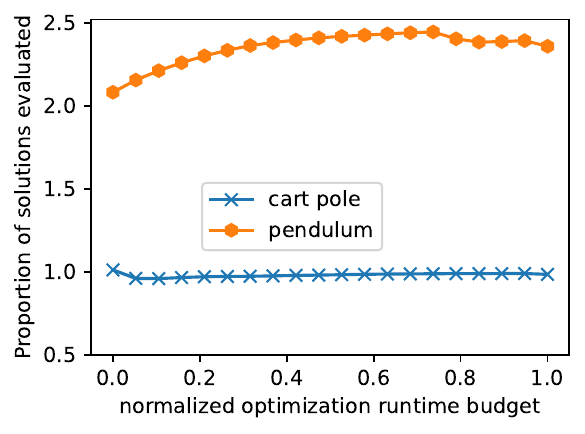}
		\caption{Ratio of solutions evaluated with and without \proposedmethodname in the same optimization time. 
		A higher value indicates that \proposedmethodname was able to evaluate more solutions in the same time.}
		\label{fig:evalsproportion_classic}
	\end{figure}

    To confirm this hypothesis, we computed the ratio of solutions evaluated with and without \proposedmethodname. 
	For instance, a ratio of $2.0$ indicates that using \proposedmethodname, twice as many solutions are evaluated in the same amount of computation time.
	The ratio of solutions evaluated for the cart-pole is 1, as shown in Figure~\ref{fig:evalsproportion_classic}, which means that the same amount of solutions are being evaluated with or without \proposedmethodname.	
	
	The results for the \textit{pendulum} task are shown in Figure~\ref{fig:gymenvnamependulum-v1expline}.
	With \proposedmethodname, a significant amount of time is saved in this task and a final higher objective value is obtained.
	For instance, with \proposedmethodname, the median time to reach an objective value of -10 is less than 150 seconds using \proposedmethodname, compared to more than 300 seconds without.
 	In addition, as shown in Figure~\ref{fig:evalsproportion_classic}, with \proposedmethodname, more than twice as many solutions are evaluated in the same amount of time, when compared with applying no early stopping.

	\subsection{Playing Super Mario (super mario)}
	\label{sec:super_mario_tasks}
	
	\citet{vermaApplyingNeuralNetworks2020} proposed learning to play the video game ``Super Mario'' released in 1985 with NEAT~\citep{stanleyEvolvingNeuralNetworks2002}.
	In this popular video game, a character Mario needs to move to the right without dying and reach the end of the level.
	In \citet{vermaApplyingNeuralNetworks2020}'s implementation, the objective function is the horizontal distance that the character has moved.

\begin{figure*}[p]

		\begin{subfigure}{0.45\textwidth}
			\includegraphics[width=\linewidth]{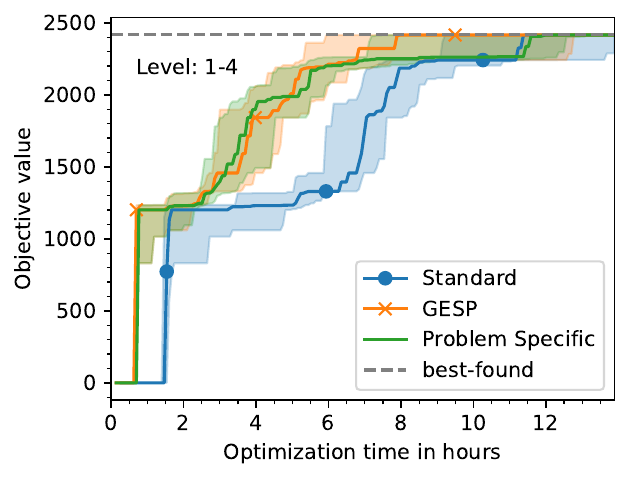}
			\label{fig:level1-4expline}
		\end{subfigure}\hfil
		\begin{subfigure}{0.45\textwidth}
			\includegraphics[width=\linewidth]{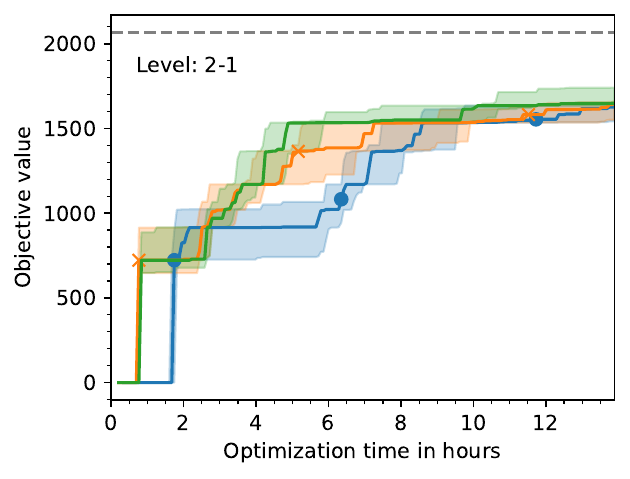}
			\label{fig:level2-1expline}
		\end{subfigure}

    \vspace{-1.2\baselineskip} 

		\begin{subfigure}{0.45\textwidth}
			\includegraphics[width=\linewidth]{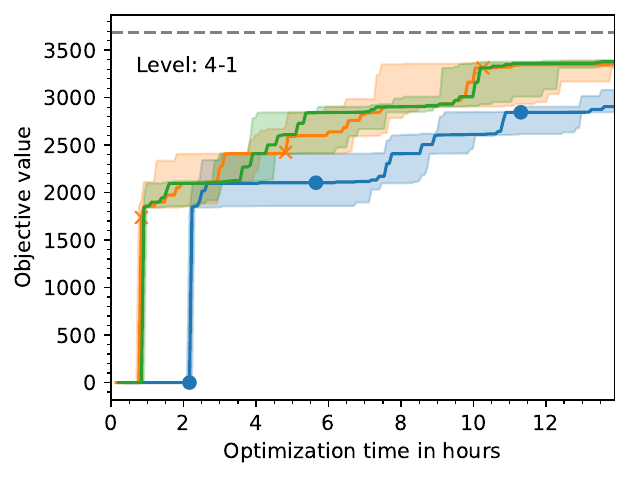}
			\label{fig:level4-1expline}
		\end{subfigure}\hfil
		\begin{subfigure}{0.45\textwidth}
			\includegraphics[width=\linewidth]{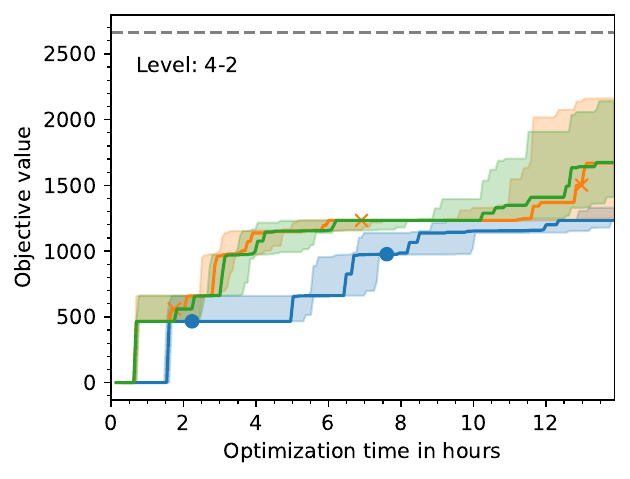}
			\label{fig:level4-2expline}
		\end{subfigure}
		
		    \vspace{-1.2\baselineskip} 
		
		\begin{subfigure}{0.45\textwidth}
			\includegraphics[width=\linewidth]{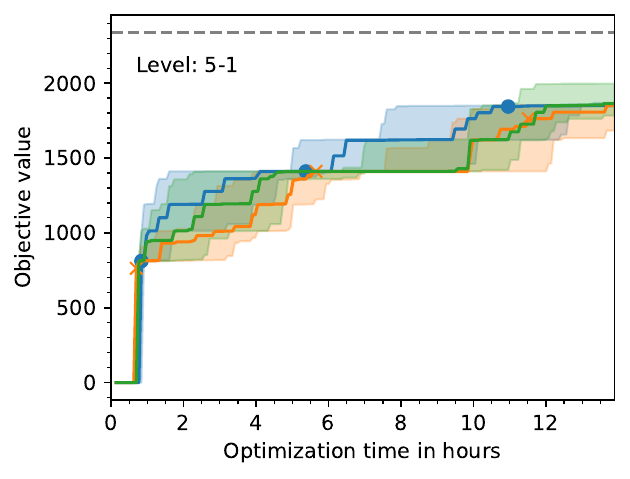}
			\label{fig:level5-1expline}
		\end{subfigure}\hfil
		\begin{subfigure}{0.45\textwidth}
			\includegraphics[width=\linewidth]{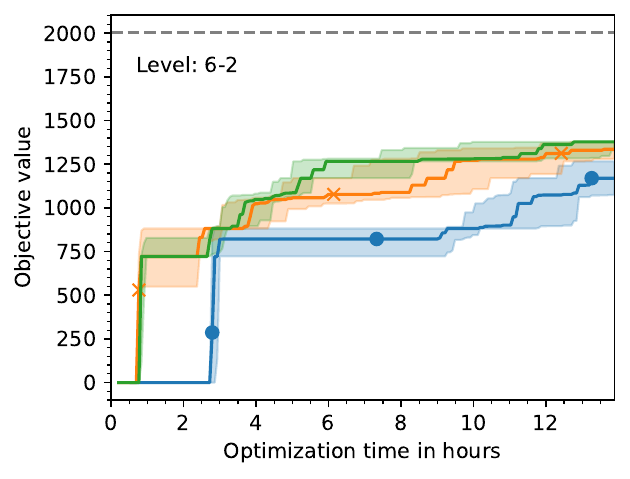}
			\label{fig:level6-2expline}
		\end{subfigure}

    \vspace{-1.2\baselineskip} 
		
		\begin{subfigure}{0.45\textwidth}
			\includegraphics[width=\linewidth]{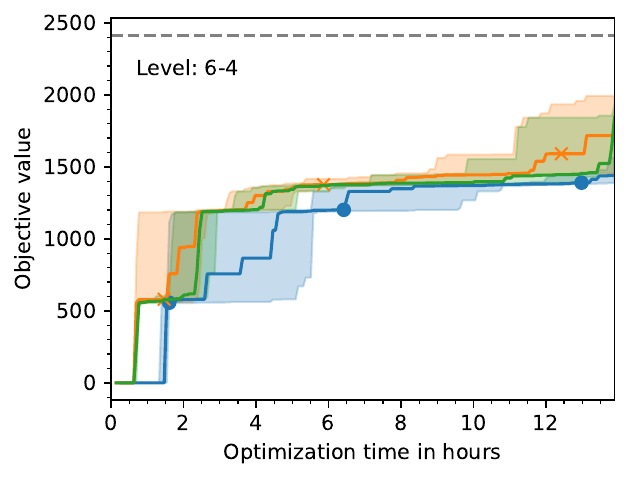}
			\label{fig:level6-4expline}
		\end{subfigure}
		    \vspace{-1.2\baselineskip} 
		\caption{The objective value of the agents with respect to computation time in \textbf{super mario} with \proposedmethodname, with the problem specific stopping criterion and without additional stopping criterion. Levels from top-left to bottom-right: 1-4, 2-1, 4-1, 4-2, 5-1, 6-2, 6-4.}
		\label{fig:results_super_mario}
		
	\end{figure*}

	A maximum episode length of 1000 steps is considered, although the episode also ends if the character dies.
	Verma implemented an additional stopping criterion: if the character does not move horizontally in 50 consecutive steps, then the episode also ends.
	In the following, we compare this problem specific stopping criterion with \proposedmethodname, and we also consider no stopping criterion as a baseline.
	We set $t_{grace} = 50$ for a fair comparison with Verma's problem specific termination criterion.
	We trained the algorithm in the levels 1-4, 2-1, 4-1, 4-2, 5-1, 6-2 and 6-4.
	We show the results in Figure~\ref{fig:results_super_mario}.

	In general, the results demonstrate that there is no big difference in performance between the problem specific method and \proposedmethodname.
	In some of the levels, the problem specific approach performs better, but this difference disappears as the computation time increases. Both methods are clearly better than using no stopping criteria.

	Curiously enough, in \href{https://web.archive.org/web/20211027022543/https://www.mariowiki.com/World_5-1_(Super_Mario_Bros.)}{level 5-1}, there is no difference between the results obtained using either of the stopping methods and the baseline method that does not use any stopping criterion.
	The reason is probably that in this level, it is hard to get stuck: there are a large number of enemies and few obstacles.
	We hypothesize that in this level, it is very easy to die, hence the execution is terminated regardless of the other stopping criteria (and therefore \proposedmethodname have no effect).

\begin{figure}
        \centering
		\includegraphics[width=0.7\linewidth]{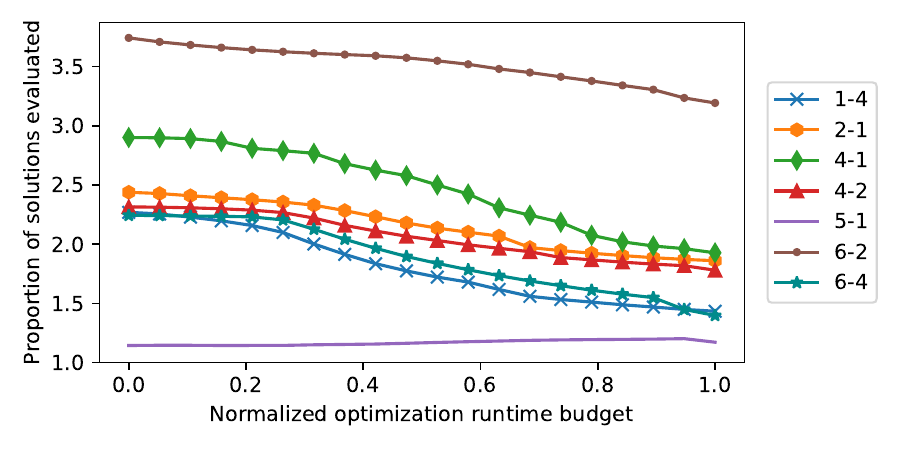}
		\caption{Ratio of solutions evaluated with and without \proposedmethodname in the same optimization time. 
		A higher value indicates that \proposedmethodname was able to evaluate more solutions in the same time.}
		\label{fig:evalsproportion_supermario}
	\end{figure}

	To validate this the hypothesis, we computed the ratio of solutions evaluated in the same optimization time with and without \proposedmethodname.
	We compute this ratio for all of the \textbf{super mario} levels, and we show the result in Figure~\ref{fig:evalsproportion_supermario}.
	As can be seen in the figure, in level 5-1 the ratio is almost $1$, indicating that \proposedmethodname rarely stops the evaluation of the agents early in this level.
	This finding validates the previous hypothesis.

	\hfil 

\FloatBarrier

	\subsection{Tasks in the Mujoco environment (mujoco)}
	Mujoco is a high performance physics simulator.
	OpenAI Gym has some \href{http://web.archive.org/web/20221006093330/https://www.gymlibrary.dev/environments/mujoco/}{tasks}\footnote{https://www.gymlibrary.dev/environments/mujoco/} defined in this environment which have been extensively used in reinforcement learning research.
    In this section of the experimentation, we experiment with the tasks \textit{half cheetah}, \textit{swimmer}, \textit{ant}, \textit{hopper}, \textit{walker2d}.
    In all of these tasks, the objective is to move the agent as far as possible from the initial position, while minimizing energy use.
    We also consider the \textit{inverted double pendulum}, in which the objective is to keep a double pendulum balanced.

    In all of these tasks, the policies are learned with CMA-ES~\citep{igelComputationalEfficientCovariance2006} (with the same algorithm that was considered in the classic control tasks in Section~\ref{sec:classic_control}) and we set the grace period parameter to 20\% of the maximum time: $t_{grace} = 0.2 \cdot t_{max}$.

\begin{figure*}
		
		\begin{subfigure}{0.45\textwidth}
			\includegraphics[width=\linewidth]{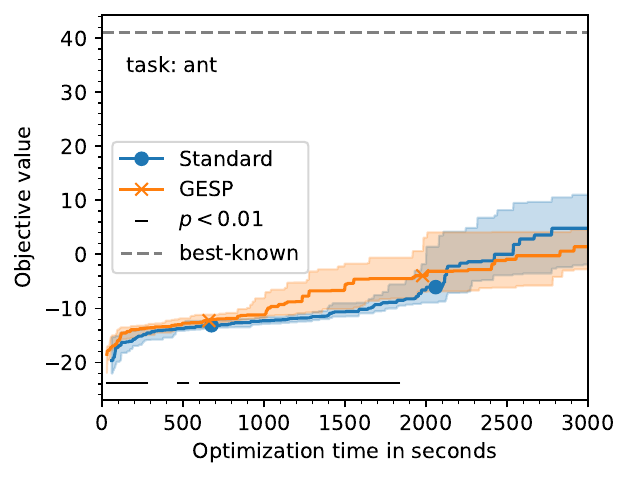}
			\label{fig:gymenvnameant-v3expline}
		\end{subfigure}\hfil
		\begin{subfigure}{0.45\textwidth}
			\includegraphics[width=\linewidth]{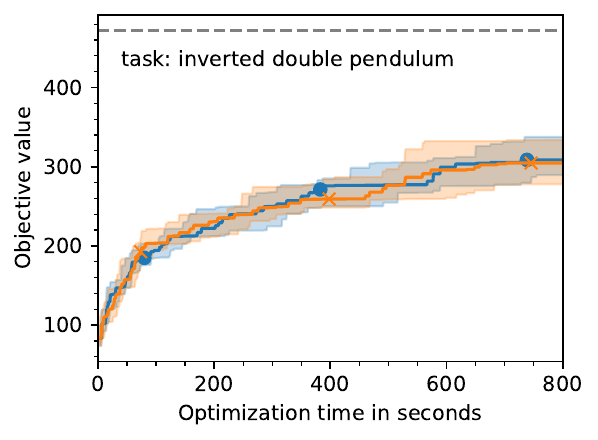}
			\label{fig:gymenvnameinverteddoublependulum-v2expline}
		\end{subfigure}
		
		\begin{subfigure}{0.45\textwidth}
			\includegraphics[width=\linewidth]{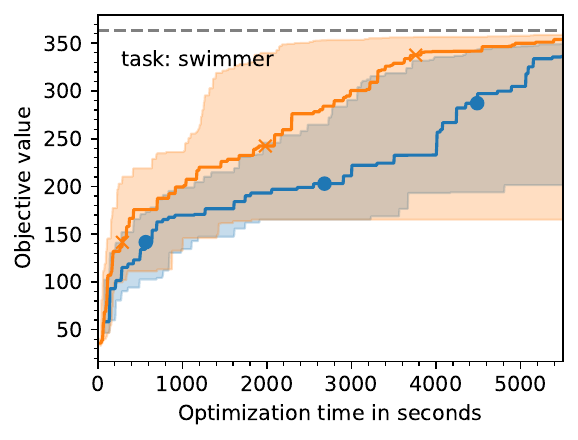}
			\label{fig:gymenvnameswimmer-v3expline}
		\end{subfigure}\hfil
		\begin{subfigure}{0.45\textwidth}
			\includegraphics[width=\linewidth]{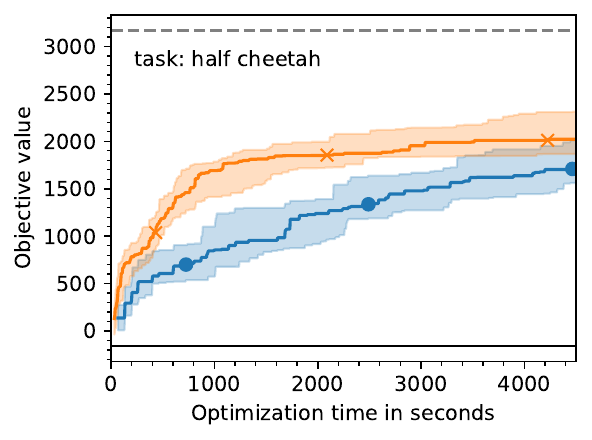}
			\label{fig:gymenvnamehalfcheetah-v3expline}
		\end{subfigure}
		
		\begin{subfigure}{0.45\textwidth}
			\includegraphics[width=\linewidth]{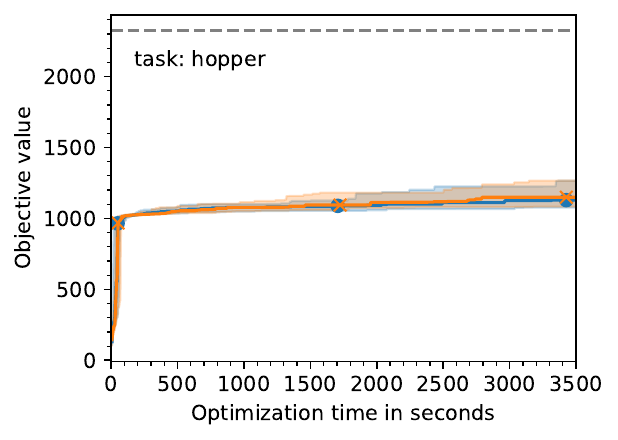}
			\label{fig:gymenvnamehopper-v3expline}
		\end{subfigure}\hfil
		\begin{subfigure}{0.45\textwidth}
			\includegraphics[width=\linewidth]{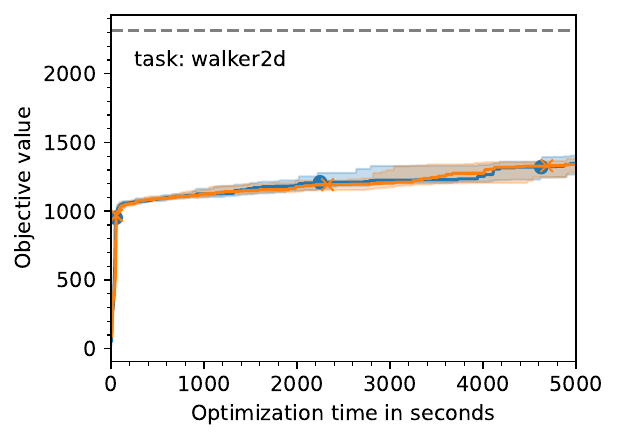}
			\label{fig:gymenvnamewalker2d-v3expline}
		\end{subfigure}
		
		\caption{The objective value of the agents with respect to computation time in the \textbf{mujoco} tasks with and without \proposedmethodname. Environments from top-left to bottom-right: \textit{ant}, \textit{inverted double pendulum}, \textit{swimmer}, \textit{half cheetah}, \textit{hopper}, \textit{walker2d}.}
		\label{fig:results_mujo_co}
		
	\end{figure*}

    The results are shown in Figure~\ref{fig:results_mujo_co}.
	In the \textit{inverted double pendulum}, \textit{hopper} and \textit{walker2d} tasks, applying \proposedmethodname made no difference in the performance and computation time.
	In the other three tasks, by using \proposedmethodname we are able to get a better objective value in the same amount of time, although the difference was only statistically significant in \textit{ant} (until 1600 seconds) and \textit{half cheetah} tasks.

	The tasks \textit{ant}, \textit{hopper}, \textit{walker2d} and \textit{inverted double pendulum} have problem specific stopping criterion that stop the evaluation when the state of the agent is `unhealthy'.
	The definition of "healthy agent" is different for each problem: for example, in the \href{http://web.archive.org/web/20220903205441/https://www.gymlibrary.dev/environments/mujoco/ant/}{\textit{ant}} task, an agent is considered healthy if all the state spaces are finite and the distance from the body of the ant to the floor is in the interval $[0.2, 1]$.
	These additional stopping criteria are essential for the learned policy to be realistic: we do not want a policy that tries to exploit the physics simulator's bugs.
	However, we hypothesize that these stopping criteria are already very good at avoiding wasting time in undesirable states, and consequently, \proposedmethodname has little room for further improvement.

\begin{figure}
	
	\begin{subfigure}{0.45\textwidth}
		\includegraphics[width=\linewidth]{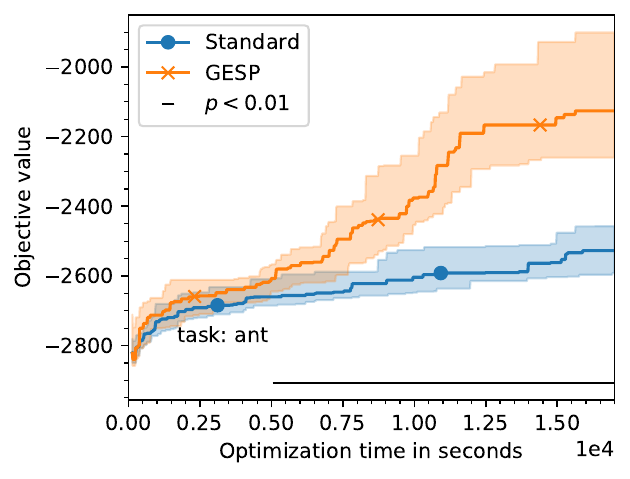}
		\label{fig:gymenvnameswimmer-v3expline_DTU}
	\end{subfigure}\hfil
	\begin{subfigure}{0.45\textwidth}
		\includegraphics[width=\linewidth]{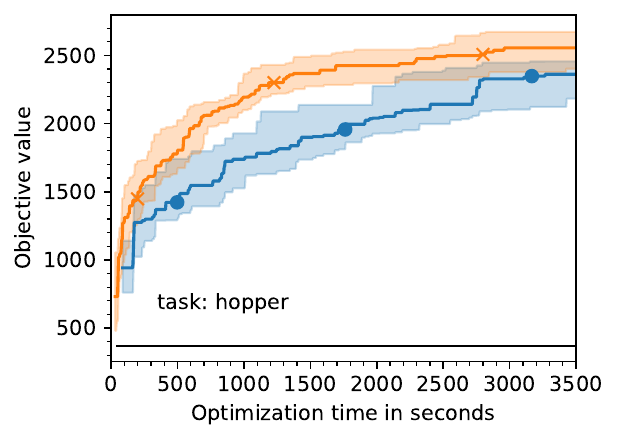}
		\label{fig:gymenvnamehopper-v3expline_DTU}
	\end{subfigure}
	
	\centering
	\begin{subfigure}{0.45\textwidth}
		\includegraphics[width=\linewidth]{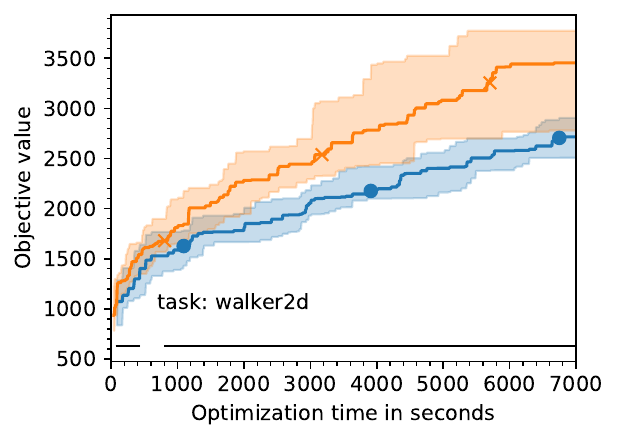}
		\label{fig:gymenvnamewalker2d-v3expline_DTU}
	\end{subfigure}
	
	\caption{The objective value of the agents with and without \proposedmethodname with respect to computation time, without stopping the evaluation when the state of the agent is unhealthy. Environments from top to bottom: \textit{ant}, \textit{hopper}, \textit{walker2d}.}
	\label{fig:results_mujo_co_DTU}
	
\end{figure}

	To validate this hypothesis, we repeated the experimentation for the tasks \textit{ant}, \textit{hopper} and \textit{walker2d} but this time without the \textit{terminate when unhealthy} stopping criterion enabled.
	We recorded the performance with respect to the optimization time with \proposedmethodname enabled and disabled.
	The results are shown in Figure~\ref{fig:results_mujo_co_DTU}.
	With the \textit{terminate when unhealthy} stopping criteria disabled, \proposedmethodname is able to save a lot of computation time in these three tasks, indicating that the method would be useful if one did not have the in-depth understanding of the task required to create problem-specific stopping criteria.
	
\begin{figure}
		\centering
		\includegraphics[width=0.9\linewidth]{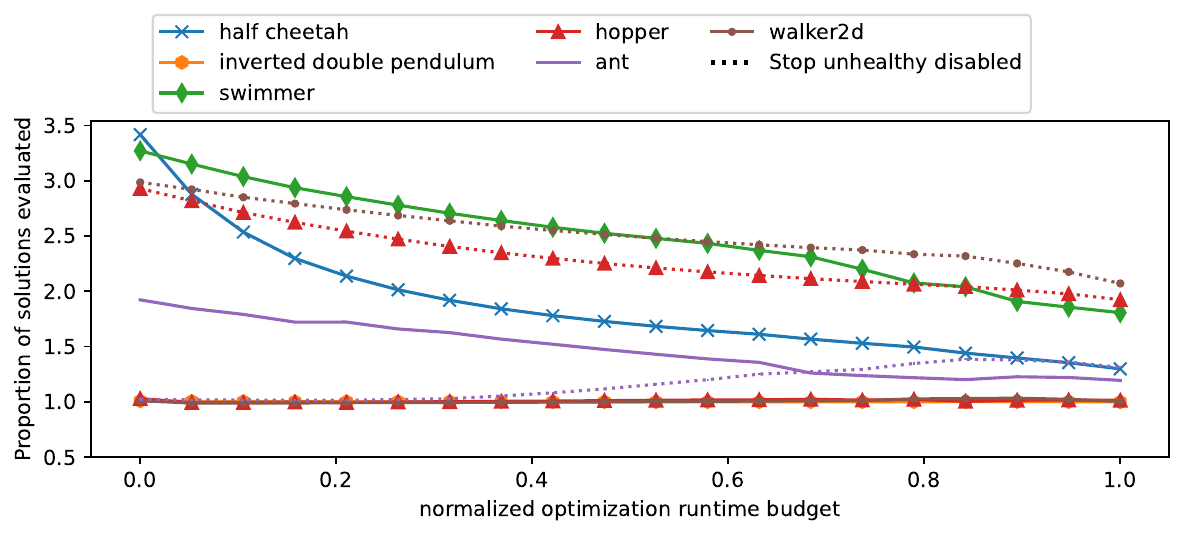}
		\caption{Ratio of solutions evaluated with and without \proposedmethodname in the same optimization time. 
			A higher value indicates that \proposedmethodname was able to evaluate more solutions in the same time.	A dashed line indicates that the results were obtained with the \textit{terminate when unhealthy} stopping criterion disabled.
		}
		\label{fig:evalsproportion_mujoco}
	\end{figure}

	We show the ratio of extra evaluations computed with \proposedmethodname in Figure~\ref{fig:evalsproportion_mujoco}.
	In the case of \textit{hopper} and \textit{walker2d}, the ratio is almost $1$, which means that \proposedmethodname is unable to save computation time in these two tasks.
	However, when we disable the \textit{terminate when unhealthy} stopping criterion, the ratio is a lot higher in these two tasks, which suggests that \proposedmethodname is able to early stop under-performing solutions.
	These results suggest that the hypothesis above is true.

	In conclusion, \proposedmethodname was able to save computation time in some of the tasks in this environment, and it is specially useful when there are no problem specific stopping criteria available.
	However, it can also save computation time alongside existing stopping criteria, although to a lesser extent (as was the case for the \textit{ant} task, as shown in Figure~\ref{fig:results_mujo_co}).

	\subsection{NIPES within the ARE framework (NIPES explore)}
	\label{sec:experiments_ARE_framework}
	Recent research in the field of Evolutionary Robotics has attempted to lay a foundation for developing frameworks that enable the autonomous design
	and evaluation of robots~\citep{eibenAutonomousRobotEvolution2021}. In contrast to much previous work in Evolutionary Robotics which typically focuses only on control, recent approaches attempt to simultaneously evolve both body and control of a robot.  For example, joint optimization of body and control  is accomplished in the framework known as ARE (Autonomous Robot Evolution)~\citep{legoffMorphoevolutionLearningUsing2021} using a nested architecture which uses an EA in an outer loop to evolve a body design and a learning algorithm within an inner loop to optimize its controller. \citet{legoffSampleTimeEfficient2020} proposed a learning algorithm to learn the control policy of wheeled robots in the inner loop dubbed NIPES for this purpose. The algorithm combines CMA-ES~\citep{igelComputationalEfficientCovariance2006} and novelty search~\citep{lehmanAbandoningObjectivesEvolution2011} in order to create a method that is a more sample and time efficient algorithm than CMA-ES alone.

\begin{figure}
		\centering
		\includegraphics[width=0.5\linewidth]{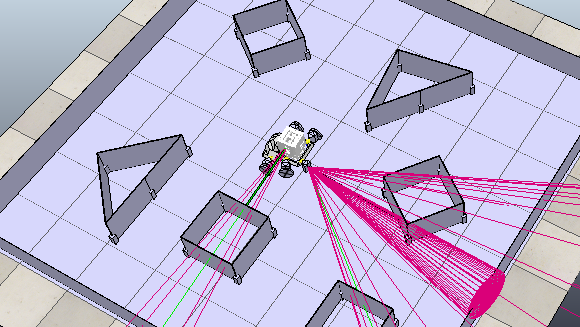}
		\caption{A screenshot of the \textit{obstacles} task.}
		\label{fig:exploration_ARE_screenshot}
	\end{figure}

    We evaluate \proposedmethodname using two exploration tasks proposed by \citet{legoffMorphoevolutionLearningUsing2021} in which a wheeled robot needs to explore an arena. 
    The goal is for the robot to explore as much of an arena as possible within 30 seconds. One arena has multiple obstacles hindering exploration, while the other has a maze-like layout that requires the robot to navigate along corridors. Each arena is divided into 64 squares (see Figure~\ref{fig:exploration_ARE_screenshot}), and the objective function is the proportion of squares visited.

	\begin{figure}
		\centering %
		
		\begin{subfigure}{0.45\textwidth}
		\includegraphics[width=\linewidth]{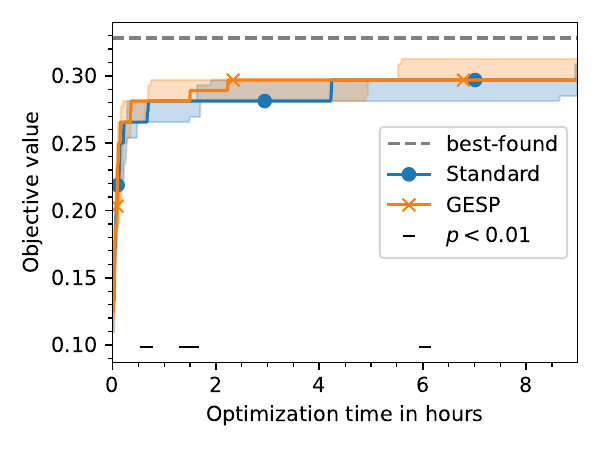}
		\caption{Obstacles}
		\label{fig:aretaskexploreobstacles}
		\end{subfigure}
		\begin{subfigure}{0.45\textwidth}
		\includegraphics[width=\linewidth]{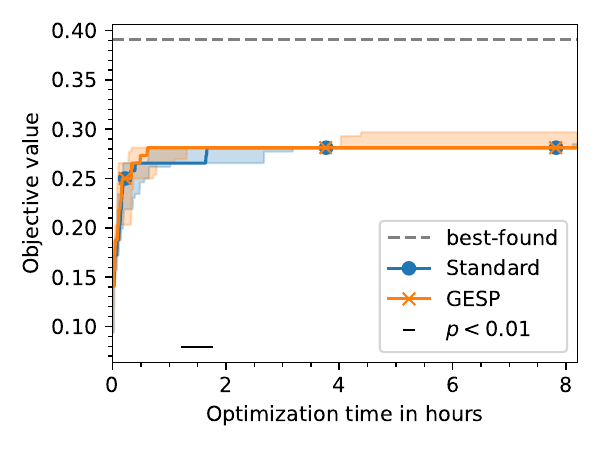}
		\caption{Hard race}
		\label{fig:aretaskexplorehard}
		\end{subfigure}
		\caption{The objective value of the agents with respect to computation time in the \textbf{NIPES explore} experiments, with and without \proposedmethodname. The black line represents that the difference is statistically significant at $\alpha= 0.01$ with a pointwise two sided Mann-Whitney test.}
		\label{fig:exploration_ARE}
	\end{figure}

	We test NIPES with and without \proposedmethodname in these two environments.
	We set the grace period parameter $t_{grace} = 0.2 \cdot t_{max}$ to 20\% of the maximum time (30 seconds as in the work by \citet{legoffMorphoevolutionLearningUsing2021}).
	The results are shown in Figure~\ref{fig:exploration_ARE}.

\begin{figure}
        \centering
		\includegraphics[width=0.5\linewidth]{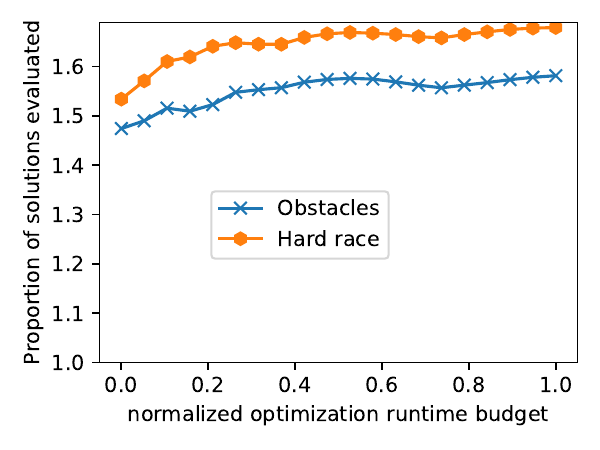}
		\caption{Ratio of solutions evaluated with and without \proposedmethodname in the same optimization time. 
		A higher value indicates that \proposedmethodname was able to evaluate more solutions in the same time.}
		\label{fig:evalsproportion_are_explore}
	\end{figure}

	In both environments (\textit{obstacles} and \textit{hard race}), \proposedmethodname improves the objective value found for the same optimization time, although the difference is not statistically significant at $\alpha = 0.01$.
    The magnitude of the difference is not observable in the figure, and by looking at the ratio of number of evaluations with or without \proposedmethodname in Figure~\ref{fig:evalsproportion_are_explore}, we can see that \proposedmethodname is able to evaluate between $40\%$ and $70\%$ more solutions in the same amount of time. This is less of a saving than in scenarios previously described (e.g. Mujoco, Super-Mario and the classic control environments).
    A higher number of repetitions (we use 30 repetitions in every experiment of this paper) might reveal a statistically significant difference, which can also be observed at $\alpha = 0.05$ (not shown in the figure).

	\FloatBarrier

	\subsection{Robotics, Evolution and Modularity (L-System)}
        \label{sec:exp_veenstra}
 
	

	Veenstra and Glette proposed a \href{https://github.com/FrankVeenstra/gym_rem2D}{simulation framework}\footnote{https://github.com/FrankVeenstra/gym\_rem2D} to evolve the morphology and control of 2D creatures~\citep{veenstra2020different} based on the OpenAI gym \href{https://web.archive.org/web/20220903205524/https://www.gymlibrary.dev/environments/box2d/bipedal_walker/}{\textit{bipedal walker}}\footnote{https://www.gymlibrary.dev/environments/box2d/bipedal\_walker/} environment.
	It is a gym environment for computationally cheap morphology search~\citep{veenstra2020different}.
	Agents start at the horizontal position $4.67$ and need to move to the right to increase their horizontal position. 	
	They propose a problem specific stopping criterion that terminates the evaluation of the current agent if its position in time $t$ is lower than or equal to $0.04 \cdot t$.
	We set the time grace parameter of \proposedmethodname to $t_{grace} = 130$, which is what the amount of frames it takes for the problem specific stopping criterion to terminate randomly generated agents.
	It is similar to the amount of frames it takes to terminate a non moving agent at $117$ frames.

\begin{figure}
		
		\begin{subfigure}[b]{0.45\textwidth}
			\includegraphics[width=\linewidth]{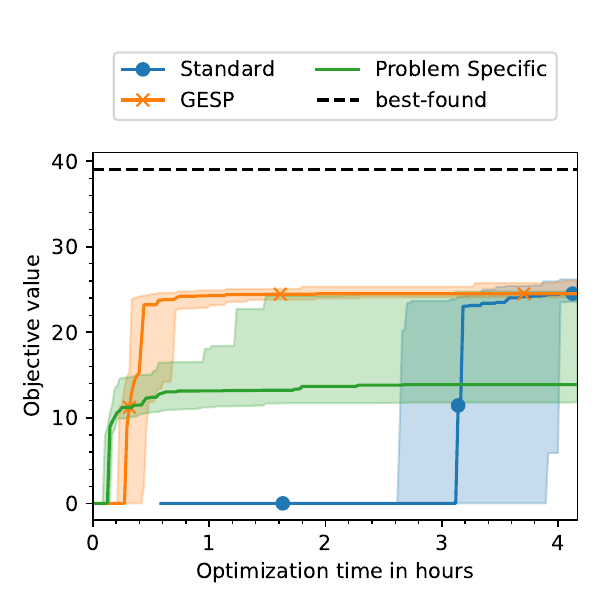}
			\caption{Objective value during training}
			\label{fig:veenstraresults}
		\end{subfigure}\hfil 
		\begin{subfigure}[b]{0.45\textwidth}
			\includegraphics[width=\linewidth]{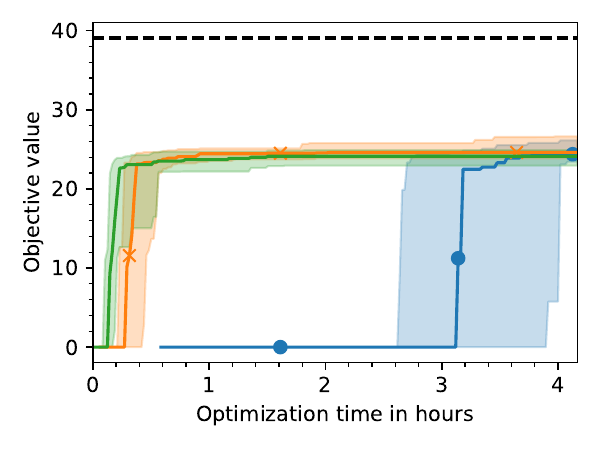}
			\caption{Objective value when reevaluating}
			\label{fig:veenstraresults_test}
		\end{subfigure}\hfil 
		
		\caption{The objective value of the agents in the task proposed \citet{veenstra2020different} (\textbf{L-System}).
			We compared \proposedmethodname, the problem specific stopping criterion, and no additional stopping criterion (Standard).
			The best found objective value is reported with respect to the total computation time.
			a) shows the best objective value observed during training, and b) shows the objective value of the best candidate in each generation when it is reevaluated with no stopping criteria.
	}
		\label{fig:veenstraresults_both}
	\end{figure}

	The results are shown in Figure~\ref{fig:veenstraresults}.
	The problem specific approach obtains a better objective value than \proposedmethodname and using no stopping criterion (Standard) at the very beginning of the optimization process.
	However, later on \proposedmethodname takes over and is better than the other two approaches until 3.6 hours.

    At the end of the training procedure, the problem specific stopping criterion obtains a poorer objective value than the other two approaches. However, we suggest that this is an artifact of the specific problem-specific stopping criteria suggested for this scenario: 
    if the best agent produced with the problem specific stopping criterion enabled is evaluated without any stopping criteria, it is possible that it might in fact obtain a better performance value. On the other hand, \proposedmethodname overcomes this limitation, as only solutions evaluated for the entire episode length (for time $t_{max}$) are candidates for the best found solution (thanks to Modification (1) introduced in Section~\ref{sec:overcoming_monotone_increasing}).

    To test whether this is the case, we repeated the experiments but reevaluating the best candidate in each generation with all the stopping criteria disabled.
    The results are shown in Figure~\ref{fig:veenstraresults_test}.
    While the performance with \proposedmethodname and Standard do not change with respect to the previous experiments, the same is not true for the problem-specific stopping criterion: in this case the problem specific approach reaches a high objective value ($>20$) slightly faster than \proposedmethodname and considerably faster than with all the stopping criteria disabled.

	The problem specific approach terminates some of the high performing agents after a while, which explains why the objective value increases more slowly for the problem specific approach without re-evaluation.
	Especially at the beginning of the optimization, solutions get terminated very quickly, because early agents move slowly.
	This makes the problem specific approach advantageous, because we waste less time on agents that can barely move, but it also means that slow moving agents that reach very far will not have a chance to be evaluated with time $t_{max}$.
	\proposedmethodname is different in that promising agents will have the chance to be evaluated until time $t_{max}$, because the purpose is to maximize the observed objective value: we are trying to solve the problem introduced in Definition~\ref{def:estimable_optimization_problem}.

	\begin{figure}
        \centering
		\includegraphics[width=0.45\linewidth]{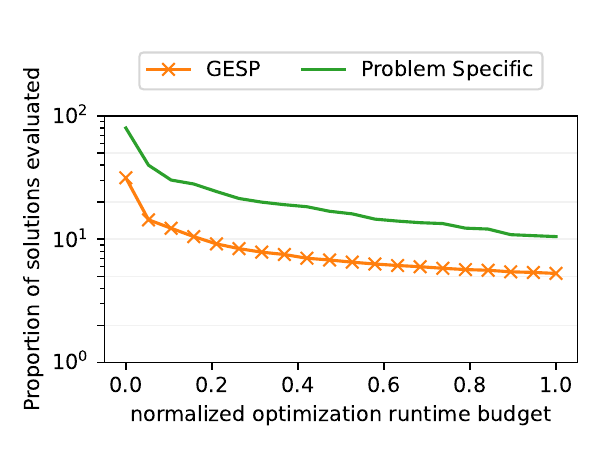}
		\caption{Ratio of solutions evaluated with and without \proposedmethodname and the problem specific stopping criterion in the same optimization time. 
		A higher value indicates that with \proposedmethodname (or the problem specific criterion), it was possible to evaluate more solutions in the same amount of time.}
		\label{fig:evalsproportion_veenstra}
	\end{figure}

    The task in this framework is to \textit{move to the right as \textbf{far} as possible}.
    However, by using the problem specific stopping criterion, agents that move slower than $0.04 \cdot t$ will eventually be terminated.
    This means that the task to be solved changes to \textit{move to the right as \textbf{fast} as possible}.

    For the purpose of the scenarios proposed in the paper by \citet{veenstra2020different}, we argue that the stopping criterion they proposed is still more suitable than using no stopping criterion or using \proposedmethodname.
    The point of their paper is to compare different encoding methods, regardless of the stopping criteria.
    By adding a problem specific stopping criterion, they significantly sped up the learning process from about three hours of computation time to 30 minutes.  
    With the problem specific approach they are able to evaluate between 10 times and 100 times more solutions in the same amount of time as shown in Figure~\ref{fig:evalsproportion_veenstra}.
    This also changed the objective function from "move as \textbf{far} as possible" to "move as \textbf{fast} as possible", however the comparison of encoding methods is also applicable to the modified version of the objective function.

    \proposedmethodname could also have been considered as the stopping criterion in their study instead of the problem specific approach.
    \proposedmethodname reduces the computation time to around 45 minutes, but unlike the problem specific approach, the objective function does not change (it is still "move as \textbf{far} as possible").


	\subsection{On the parameter $t_{grace}$}
	\label{section:tgrace_experimentatino}

	As previously explained, \proposedmethodname proposes to stop the evaluation only after $t_{grace}$ steps have been computed ($t > t_{grace}$) and Equation~\ref{eq:early_stopping_generalized} is satisfied.
	In this sense, the $t_{grace}$ parameter controls how early will evaluations be stopped.
	A lower value of $t_{grace}$ allows more time to be saved (it allows the the evaluations to be stopped earlier), but it also increases the probability of terminating good performing agents that observe a momentary low reward during their evaluation.
	If we set $t_{grace} = 0$, then any candidate solution will get discarded as soon as it does worse than the best found solution in any time step.
	Inversely, if we set $t_{grace}$ to $100\%$ of the maximum episode length, then no solutions will be terminated early, and this is equivalent to not applying \proposedmethodname (we use ``Standard'' to refer to this from now on).

	Hence, there is a trade-off between the amount of computation time that can be saved vs. how likely it is that the evaluation of a good performing agent will be terminated early.
	In this section, we carried out two experiments to study the effect of the $t_{grace}$ parameter.
	In a first experiment, we compare different $t_{grace}$ values to see which gives the best results.
	Then, in a second experiment, we measure three interesting properties with respect to the $t_{grace}$ parameter.
	For example, we measure the probability of early stopping (and therefore missing) a new best found solution, which decreases as $t_{grace}$ increases.

	\subsubsection{Performance with respect to $t_{grace}$}
	\label{sec:tgrace_different_values}
	
\begin{figure}
	\centering

	\begin{subfigure}[b]{0.49\textwidth}
		\includegraphics[width=\textwidth]{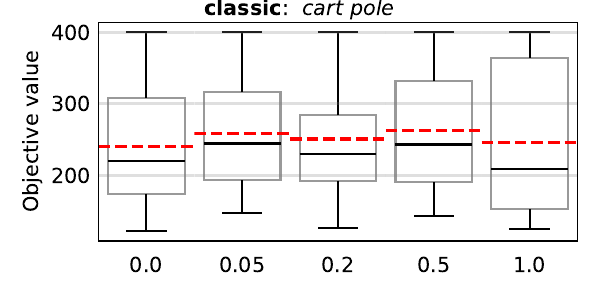}
	\end{subfigure}
	\hfill 
	\begin{subfigure}[b]{0.49\textwidth}
		\includegraphics[width=\textwidth]{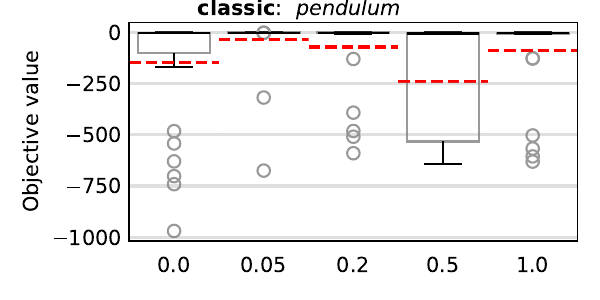}
	\end{subfigure}

	\begin{subfigure}[b]{0.49\textwidth}
		\includegraphics[width=\textwidth]{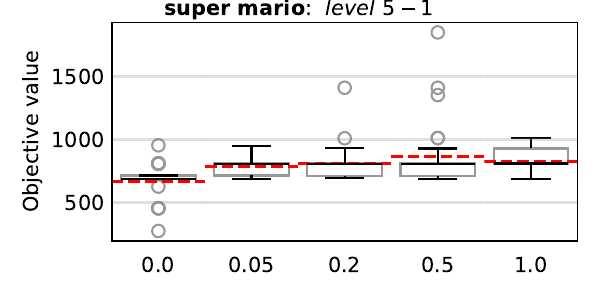}
	\end{subfigure}
	\hfill 
	\begin{subfigure}[b]{0.49\textwidth}
		\includegraphics[width=\textwidth]{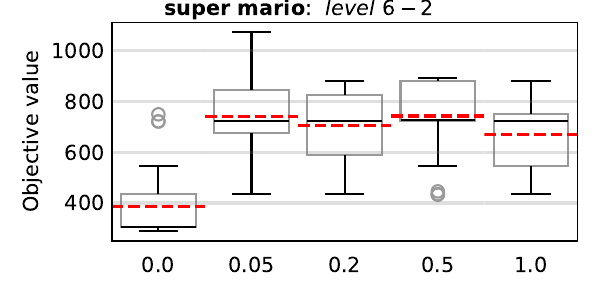}
	\end{subfigure}

	\begin{subfigure}[b]{0.49\textwidth}
		\includegraphics[width=\textwidth]{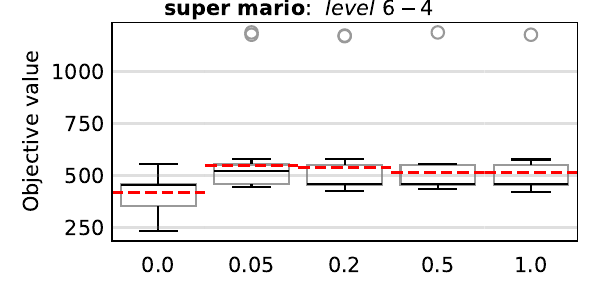}
	\end{subfigure}
	\hfill 
	\begin{subfigure}[b]{0.49\textwidth}
		\includegraphics[width=\textwidth]{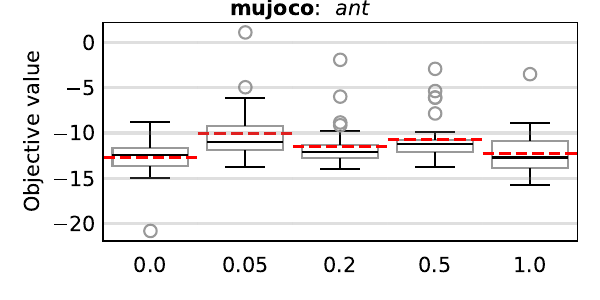}
	\end{subfigure}

	\begin{subfigure}[b]{0.49\textwidth}
		\includegraphics[width=\textwidth]{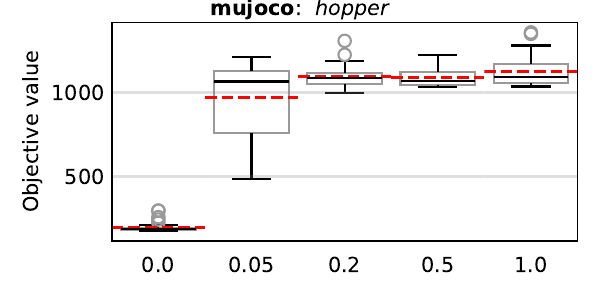}
	\end{subfigure}
	\hfill 
	\begin{subfigure}[b]{0.49\textwidth}
		\includegraphics[width=\textwidth]{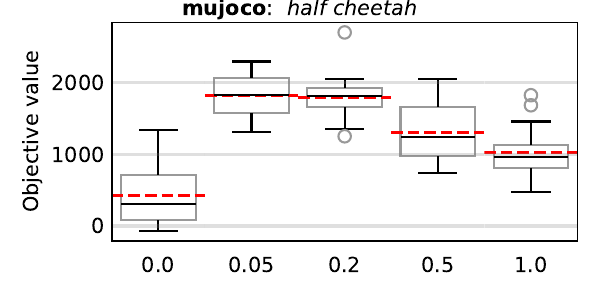}
	\end{subfigure}
	
	\begin{subfigure}[b]{0.49\textwidth}
		\includegraphics[width=\textwidth]{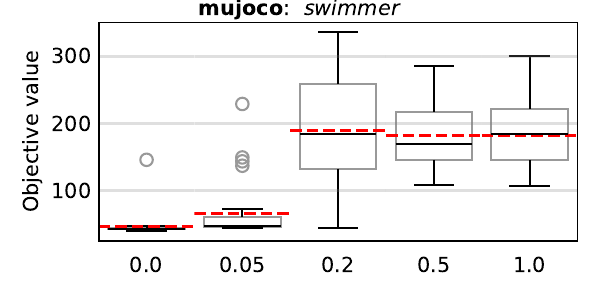}
	\end{subfigure}
		\hfill 
	\begin{subfigure}[b]{0.49\textwidth}
		\includegraphics[width=\textwidth]{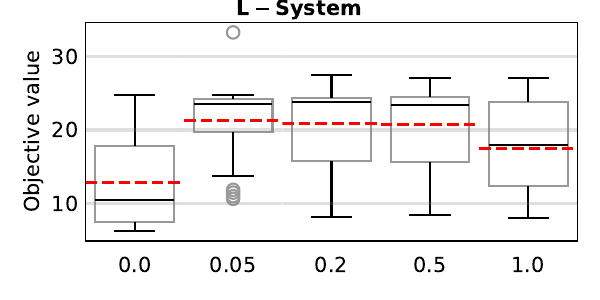}
	\end{subfigure}

	\caption{
		Box-plots showing the performance with $t_{grace}$ set to $0, 0.05, 0.2, 0.5$ and $1.0$ times \( t_{max} \).
		The red dashed horizontal line represents the average, and the black line inside the box is the median (higher is better).
		Note that setting $t_{grace} = 1.0 \cdot t_{max}$ is equivalent to not applying \proposedmethodname (Standard), as no solution is stopped early with this value of the parameter.
	}
	\label{fig:tgrace_different_values_f}
\end{figure}

	In this first experiment, we compare different values of the $t_{grace}$ parameter.
	To this end, we run \proposedmethodname with $t_{grace}$ set to $0, 0.05, 0.2, 0.5$ and $1.0$ times the maximum episode length.
	We record the best objective value observed on a subset of the tasks in the previous section.
	The experiment is carried out on three \textbf{super mario} levels, the two \textbf{classic control} tasks, four \textbf{mujoco} tasks and the \textbf{L-systems} task.
	We repeat the experiment 30 times for each value of $t_{grace}$, with the boxplots of the results shown in Figure~\ref{fig:tgrace_different_values_f}.

	On the \textit{cart pole} task, all of the parameter values perform roughly the same.
	This is the expected result, as applying \proposedmethodname has no effect and is equivalent\footnote{By definition, \proposedmethodname never stops the evaluation early on \textit{cart pole}, as explained on Section~\ref{sec:classic_control}.} to Standard (we use Standard to refer to using no early stopping, which is equivalent to setting $t_{grace} = 1.0 \cdot t_{max}$, or the rightmost boxplot in each task on Figure~\ref{fig:tgrace_different_values_f}).

	On the rest of the tasks, we cannot say that a parameter value is better than the rest, as the performance varies across tasks.
	However, applying \proposedmethodname with $t_{grace} = 0.0 \cdot t_{max}$ performs (on average) worse than Standard in the rest of the tasks.
	This suggests that the parameter $t_{grace}$ should always be set to more than $0.0 \cdot t_{max}$.

	On the \textit{pendulum} task, two of the \textbf{super mario} tasks and two of the \textbf{mujoco} tasks; the best performance is observed with $t_{grace} = 0.05 \cdot t_{max}$.
	However, $t_{grace} = 0.05 \cdot t_{max}$ also obtains a worse performance than Standard on three of the tasks: \textbf{super mario}: \textit{level 5-1} and \textbf{mujoco}: \textit{swimmer} and \textit{hopper}.
	On Section~\ref{sec:super_mario_tasks}, we observed that \proposedmethodname offered no improvement over Standard on \textbf{super mario}: \textit{level 5-1}.
	The reason is that there are a lot of enemies on \textit{level 5-1} that often kill Mario, triggering the problem specific stopping criterion, and hence there is no need to add additional stopping criteria.
	Consequently, \proposedmethodname with $t_{grace} = 0.05 \cdot t_{max}$ might be stopping the evaluation of promising candidates early on \textbf{super mario}: \textit{level 5-1}, while offering no significant time savings.
	\proposedmethodname with $t_{grace} = 0.05 \cdot t_{max}$ provides a huge drop in performance on \textbf{mujoco}: \textit{swimmer}, which can also be explained by the low probability of not missing a new best solution (studied in the second part of the Experimentation on the $t_{grace}$ parameter, on Section~\ref{sec:tgrace_nokill}).

	\proposedmethodname $t_{grace} = 0.2 \cdot t_{max}$ offers the best performance in general, while avoiding the big drops in performance on some of the tasks.
	While $t_{grace} = 0.05 \cdot t_{max}$ can sometimes speed up the optimization process more than $t_{grace} = 0.2 \cdot t_{max}$, the latter does not worsen the observed objective value in any of the tasks.
	Therefore, and based on these results, we recommend $t_{grace} = 0.2 \cdot t_{max}$ as the default parameter.

	To sum up, in this experiment we observed that:
	
	\begin{itemize}
		\item Setting \( t_{grace} \) to \( 0.2 \cdot t_{max} \) provides the optimal balance (among the tested configurations) between maximizing the optimization time saved without compromising the performance in some of the tasks.
		\item The parameter value $t_{grace} = 0.0 \cdot t_{max}$ provides a poor performance.
		\item The optimal value for the parameter $t_{grace}$ varies across tasks.
	\end{itemize}

\subsubsection{A closer look on $t_{grace}$}
\label{sec:tgrace_nokill}

To further study the $t_{grace}$ parameters, we conduct another experiment.
On the same tasks as in the previous experiment, we run the direct policy search algorithms 30 times per task with no problem specific stopping criterion, and we record the objective value observed at \textit{each} time step.
Then, we analyze these observed partial objective values by applying early stopping with \proposedmethodname for different values of $t_{grace}$, assuming that the search trajectory does not change.
Although this assumption is not as realistic as the previous experiment, it allows us to study the behaviour of \proposedmethodname changes with respect to $t_{grace}$.

Specifically, we measure three interesting proportions when $t_{grace}$ is set from $0.0$ to $1.0$ times \( t_{max} \).
First we compute the proportion of \textit{best solution not missed}, which is the probability that the best found solution with \proposedmethodname is the same as without it.
We also measure \textit{steps computed}, which measures the efficiency of $t_{grace}$ in terms of the ratio of steps computed with and without \proposedmethodname.
For instance, a \textit{steps computed} of 0.7 implies that, on average, the number of steps computed per policy was $30\%$ lower with \proposedmethodname.
Finally, we recorded \textit{\proposedmethodname improves result}, which is an upper bound\footnote{It is an upper bound because in this experiment, we assume that the search trajectory (the solutions that are visited) are the same with and without \proposedmethodname, which is not true. With \proposedmethodname, the policies are not evaluated completely, hence the search procedure is less efficient.} of the probability that with \proposedmethodname the final result is equal or better than without it.

\begin{figure}
	\centering
	
	\begin{subfigure}[b]{0.478\textwidth}
		\includegraphics[width=\textwidth]{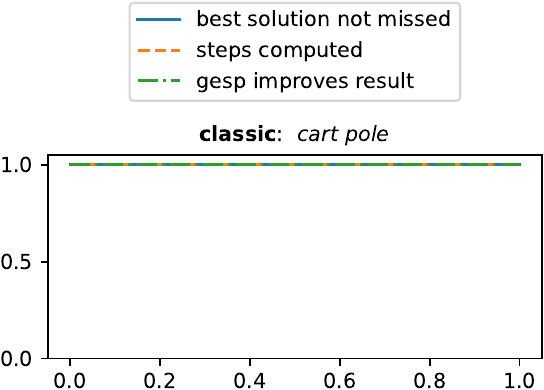}
	\end{subfigure}
	\hfill 
	\begin{subfigure}[b]{0.49\textwidth}
		\includegraphics[width=\textwidth]{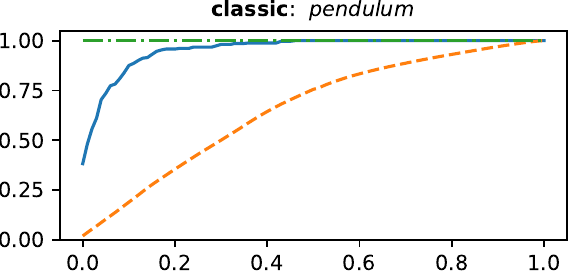}
	\end{subfigure}
	\vspace{0.5em}
	
	\begin{subfigure}[b]{0.49\textwidth}
		\includegraphics[width=\textwidth]{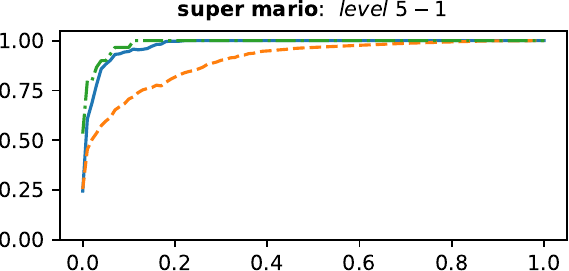}
	\end{subfigure}
	\hfill 
	\begin{subfigure}[b]{0.49\textwidth}
		\includegraphics[width=\textwidth]{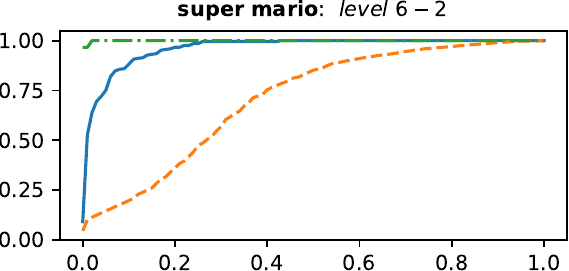}
	\end{subfigure}
	\vspace{0.5em}
	
	\begin{subfigure}[b]{0.49\textwidth}
		\includegraphics[width=\textwidth]{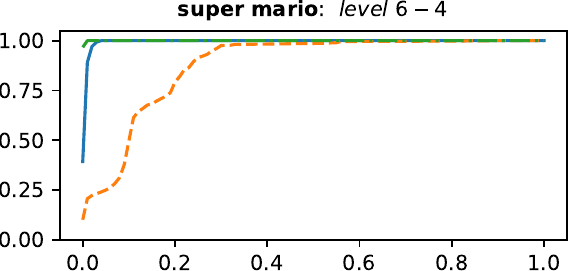}
	\end{subfigure}
	\hfill 
	\begin{subfigure}[b]{0.49\textwidth}
		\includegraphics[width=\textwidth]{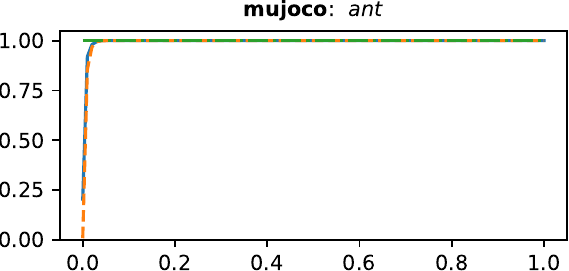}
	\end{subfigure}
	\vspace{0.5em}
	
	\begin{subfigure}[b]{0.49\textwidth}
		\includegraphics[width=\textwidth]{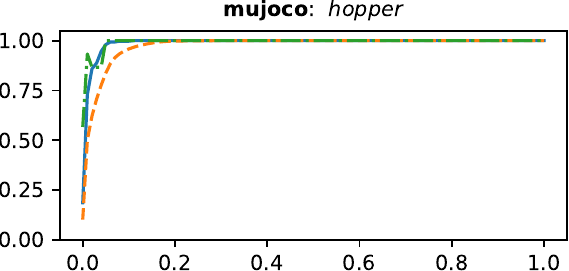}
	\end{subfigure}
	\hfill 
	\begin{subfigure}[b]{0.49\textwidth}
		\includegraphics[width=\textwidth]{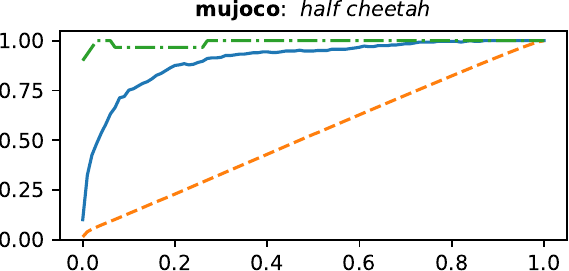}
	\end{subfigure}
	\vspace{0.5em}

	\begin{subfigure}[b]{0.49\textwidth}
		\includegraphics[width=\textwidth]{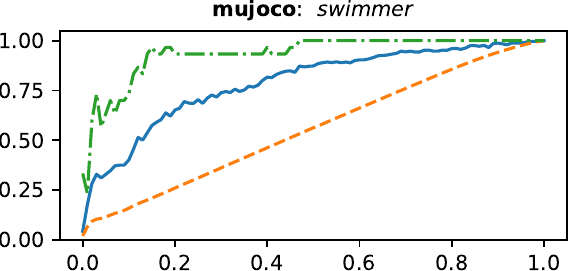}
	\end{subfigure}
	\hfill 
	\begin{subfigure}[b]{0.49\textwidth}
		\includegraphics[width=\textwidth]{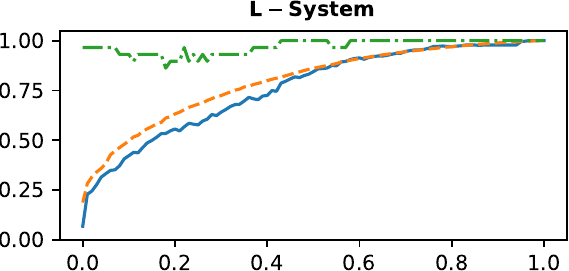}
	\end{subfigure}
    \vspace{-0.5\baselineskip} 
	\caption{
	Proportions of \textit{best solution not missed}, \textit{steps computed}, and \textit{\proposedmethodname improves result} as a function of \( t_{grace} / t_{max}\) on different tasks. 
	The x-axis represents $t_{grace}$ as a fraction of $t_{max}$, where fore example a value of 0.3 implies \( t_{grace} = 0.3 \cdot t_{max} \).
	}
	\label{fig:tgrace_nokill_proportions}
\end{figure}

The results are shown in Figure~\ref{fig:tgrace_nokill_proportions}.
All the proportions are constantly $1.0$ in \textit{cartpole}, as applying \proposedmethodname has no effect. 
For the rest of the tasks, the \textit{best solution not missed} proportion starts low, and quickly increases as $t_{grace}$ increases.
Except on the two lasts tasks, a value of $t_{grace} = 0.2 \cdot t_{max}$ is enough for this proportion to be higher than 0.8, which means that in general, setting $t_{grace} = 0.2 \cdot t_{max}$ has associated a high probability of not missing the best solution.
In contrast, the proportion of \textit{steps computed} raises more slowly with $t_{grace}$.

Hence, setting $t_{grace} = 0.2 \cdot t_{max}$ has associated a high proportion of \textit{best solution not missed}, while \textit{steps computed} is as low as possible.
In contrast, with a value of $t_{grace} = 0.0$, it is very likely that the best found solution will be missed with \proposedmethodname, which explains why it was the worst parameter setting in the experiment in the previous section.

Regarding \textit{\proposedmethodname improves result}, it is generally high, but again this is an upper bound of the probability that \proposedmethodname improves the result.
In \textit{swimmer}, \textit{\proposedmethodname improves result} is very low when $t_{grace}$ is set to $0.0$ and $0.05$ times $t_{max}$, which explains why these parameter values performed so poorly in the previous experiment.

	In short, in this experiment we saw that:

\begin{itemize}
	\item Setting \( t_{grace} \) to \( 0.2 \cdot t_{max} \) provides a good balance between maximizing the optimization time saved without compromising the probability of finding the same best solution.
	\item The parameter value $t_{grace} = 0.0$ is likely to miss the best found solution.
	\item The probability of \textit{best solution not missed} and the proportion of \textit{steps computed} go to $1.0$ as we increase \( t_{grace} \), but the former increases much faster.
\end{itemize}

	\subsection{Discussion and future work}
	\label{section:discussion_and_future_work}

	In the previous sections, we experimented with \proposedmethodname in \nframeworksexperimentation different direct policy search environments (see Table~\ref{tab:summary_results} for a summary of the experimentation).
	\proposedmethodname maintained or improved the performance of the candidate solutions trained for the same amount of computation time in all the tasks considered in the paper.
	In general, the biggest improvement with \proposedmethodname was observed when \proposedmethodname early stopped the evaluation of poorly performing candidates.

    \begin{table}
    	\centering
    	\resizebox{\linewidth}{!}{%
    		\begin{tabular}{>{\centering\hspace{0pt}}m{0.15\linewidth}>{\centering\hspace{0pt}}m{0.14\linewidth}>{\centering\hspace{0pt}}m{0.14\linewidth}>{\hspace{0pt}}m{0.65\linewidth}}
    			                           & \multicolumn{2}{c}{speedup with \proposedmethodname\unskip*}                    & Additional Conclusions \\ \hline
                                              & first half                 & last half                    &                                                                                                                              \\ \hline

    			\textbf{classic control}  &   1 out of 2     &    1 out of 2                  & \proposedmethodname has no effect in \textit{cart pole}. \proposedmethodname works in \textit{pendulum} even though it has a monotone decreasing objective function.    \\ \hline
                \textbf{super mario}      &   6 out of 7     &    4 out of 7                  & \proposedmethodname makes no improvement when the environment itself stops the evaluation early (mario touches an enemy and dies). \\ \hline
                \textbf{mujoco}           &   2 out of 5     &    1 out of 5                  & \proposedmethodname generates additional speedup when the problem specific stopping criteria are disabled. \\ \hline
                \textbf{NIPES explore}    &   0 out of 2     &    0 out of 2                  & The speedup with \proposedmethodname is small in magnitude, and is not statistically significant at $\alpha=0.01$. \\ \hline
    			\textbf{~~ L\nobreakdash-System ~~}  &   1 out of 1     &    1 out of 1       &  Unlike the problem specific stopping criterion, \proposedmethodname does not change the definition of the objective function. \\ \hline
                & & & \\
               \multicolumn{4}{l}{*Number of scenarios in which \proposedmethodname obtained a better score in the first/last half of the} \\
               \multicolumn{4}{l}{optimization process for the same amount of computation time.} \\
      
      \end{tabular}%
      }
    	\caption{Summary of the experimental results.}
    	\label{tab:summary_results}
    \end{table}

	However, in some tasks, there was no improvement when applying \proposedmethodname.
	When other terminating criteria already stopped the evaluation, \proposedmethodname produced no further improvement.
	We observed this for level 5-1 in \textbf{super mario} and for \textit{walker2d} and \textit{hopper} in \textbf{mujoco}, where \proposedmethodname provided no benefit but also did not negatively impact the performance.

    We have shown that applying \proposedmethodname to direct policy search is generally beneficial.
    Firstly, in the experimentation carried out, \proposedmethodname never made the results worse: in the worst case, it made no difference.
    In addition, unlike problem specific approaches, it does not require problem specific knowledge and is simpler to implement than other approaches such as surrogate models.
    Moreover, it does not change the objective function (unlike for example the problem specific stopping criterion in \textbf{L-System}).


    \paragraph{Beyond direct policy search}

	The reinforcement learning algorithms considered in this paper carry out \textit{direct policy search through evolutionary algorithms}.
	These algorithms are simple to implement and understand how they work.
	They do not require value function estimations, and because of their simplicity, are more unlikely to suffer from unstability issues.
	Despite the advantages of direct policy search methods in terms of ease of implementation, understanding and low computation cost, their performance is much lower than other more elaborate techniques that use value estimation, or even other policy learning methods such as Proximal Policy Optimization~\citep{schulman2017proximal}.
	These more advanced techniques consider larger policy spaces and more complex algorithms to find better policies that can be achieved with direct policy search.

	However, we argue that the purpose of this paper is not to maximize the performance of the agents in the environments.
	Instead, the purpose of this paper is to introduce a general early stopping for direct policy search that requires no problem specific information and is applicable in many environments.
	For that matter, we chose existing direct policy search approaches in the literature, and tried to show the benefit that early stopping can bring to these policy learning algorithms.
	As direct policy search methods are much simpler than other state of the art approaches, the performance difference with respect to state of the art reinforcement learning algorithms is large.
	As future work, the proposed methodology could be adapted to other more sophisticated episodic learning approaches beyond \textit{direct policy search through evolutionary algorithms}.


\paragraph{Constraints} 
The proposed methodology is in general applicable even when constraints are considered. In the case of \textit{A priori} constraints~\citep{digabelTaxonomyConstraintsSimulationBased2015} (checking for the feasibility of a solution requires no additional computation), \proposedmethodname can be applied only to feasible solutions, after the feasibility check, with no additional changes required.
When the constraint is simulation based~\citep{digabelTaxonomyConstraintsSimulationBased2015}, the constraint can be thought of as a problem specific stopping criterion: when the solution is found to be unfeasible, the evaluation is stopped.

An example of such a constraint is present in the \textit{ant} mujoco task, where the goal is for an ant shaped robot to move as far as possible from the initial position.
In this task, the evaluation of a solution is terminated when any of the state variables of the ant are not correctly defined, or torso and the floor is not inside the interval $[0.2,1.0]$, which are essentially feasibility checks.
The experimentation with \proposedmethodname also considers this \textit{ant} environment with the mentioned constraint, and in short, we observed that \proposedmethodname can work alongside these kinds of problem specific stopping criteria.

	\paragraph{Deceptive reward}

	Finally, it remains to be seen whether the method can be applicable on tasks in which there is a deceptive reward, i.e. when the reward may deceptively encourage the agent to perform actions that prevent it from discovering the globally optimal behaviour, leading to convergence to a local optimum.
	Deceptive rewards are challenging in evolutionary robotics in general, because they can lead to premature convergence~\cite{doncieuxBlackboxOptimizationReview2014}. 
	In such cases, a policy search algorithm must be able to select solutions with a low reward to be able to eventually reach the best solutions. 
	A good example of such task is maze-solving (e.g. the hard maze used in the work of \citet{lehmanAbandoningObjectivesEvolution2011}) in which reward is often measured as a Euclidean distance from the end-point.
	On such tasks, early stopping methods (like \proposedmethodname) are unlikely to work well as a poor reward can lead to early termination.

	The field of multi-fidelity optimization features a similar challenge to the deceptive rewards problem. 
	When the fidelity of the model is too low, the optimization algorithm can converge to a solution that is optimal on low fidelity models, but it is not optimal on the real objective function~\citep{liu2016multi-fidelity,robinson2006multifidelity,alizadeh2020managing,7106543}. 
	This is known as false convergence, and analogous to learning a policy that is too greedy because early stopping was used.

	Even though \proposedmethodname is unlikely to work in direct policy search with deceptive rewards, it might still be interesting to test it to gain more understanding with respect to how the method might be adapted in future to cope with this kind of reward.
	In addition, there are other potential changes that might improve the applicability of \proposedmethodname in these settings. 
	For example, novelty search~\citep{lehmanAbandoningObjectivesEvolution2011} has shown promising results in problems with deceptive rewards, and \proposedmethodname could be adapted in this context such that candidate solutions that do not show novel behaviour (and also have poor performance) are terminated early.

	\proposedmethodname is a very simple stopping criterion, in which the evaluation is stopped when it performs worse than the best found solution with extra $t_{grace}$ evaluation time. 
	The contribution of this work is to show that general early stopping is applicable to direct policy search through a simple to implement method.
	However, it might also be possible to consider other more sophisticated early stopping criteria by aggregating several evaluations, as in \citet{desouzaCappingMethodsAutomatic2022}'s work.
	Although they are not directly compatible with direct policy search problems, as future work, it would be interesting to adapt \citet{desouzaCappingMethodsAutomatic2022}'s methods for policy learning.
	Comparing different general early stopping criteria for direct policy search is also an interesting idea for future work.

	\section{Conclusion}
	\label{section:conclusion}
	
	In this paper, we introduced \proposedmethodname: an early stopping method for optimization problems suitable to both increasing and decreasing objective functions.
	Unlike problem specific early stopping criteria, \proposedmethodname is general and applicable to many problems: it does not use domain specific knowledge.

	In a wide ranging set of experiments, we showed that adding \proposedmethodname as an additional stopping criterion usually saves a significant amount of computation time in direct policy search tasks, and allows a better objective value to be found in the same computation time.
    Moreover, \proposedmethodname did not decrease the objective value in any of the tested environments.
	We also compared \proposedmethodname to problem specific stopping criteria, and concluded that in general, \proposedmethodname had a similar performance to problem specific approaches while being more generally applicable.

	We do not claim that \proposedmethodname is better than problem specific approaches. 
    Neither do we propose that \proposedmethodname is a substitute to them.
	They both have strengths: problem specific approaches can exploit domain knowledge, because the researcher implementing them might have insight into when an agent is wasting time.
	\proposedmethodname, on the other hand, does not require domain knowledge and is applicable `out of the box' to many problems.
	We argue that \proposedmethodname is a useful early stopping mechanism applicable to problems that have no problem specific early stopping approaches.
	Moreover, it can also be introduced in addition to problem specific approaches.	\proposedmethodname is most useful for optimization problems that have costly and lengthy function evaluations. 
    We have evaluated the method in the context of a number of classic control problems and in a robotics domain, however, there are obvious opportunities to extend the approach to other domains which have an expensive objective function, for example optimization of production processes~\citep{chen2021efficient}.
 
    Code to reproduce all the experiments in the paper together with a brief explanation on how to apply the method are available in a GitHub repository \href{https://github.com/EtorArza/GESP}{https://github.com/EtorArza/GESP}.
 

	
	\textbf{Acknowledgments:}


Etor Arza is partially supported by the Basque Government through the BERC 2022-2025 and the ELKARTEK program KONFLOT KK-2022/00100 and by the Spanish Ministry of Economy and Competitiveness, through BCAM Severo Ochoa excellence accreditation SEV-2023-2026 and the research project PID2019-106453GA-I00/AEI/10.13039/501100011033.  Emma Hart and Léni Le Goff are supported by the Edinburgh Napier University through EPSRC EP/R035733/1.

	
	\bibliographystyle{plainnat}
	\bibliography{main}

\end{document}